\documentclass[lettersize,journal]{IEEEtran}
\usepackage{amsmath,amsfonts}
\usepackage{algorithmic}
\usepackage{algorithm}
\usepackage{array}
\usepackage[caption=false,font=normalsize,labelfont=sf,textfont=sf]{subfig}
\usepackage{textcomp}
\usepackage{stfloats}
\usepackage{url}
\usepackage{verbatim}
\usepackage{graphicx}
\usepackage{cite}
\usepackage[dvipsnames]{xcolor}
\usepackage{multirow}
\usepackage[inkscapeformat=png]{svg}
\usepackage{kotex}
\usepackage{booktabs}

\newcommand{\nj}[1]{\textcolor{black}{#1}}
\newcommand{\sy}[1]{\textcolor{black}{#1}}
\newcommand{\yr}[1]{\textcolor{black}{#1}}
\newcommand{\yyr}[1]{\textcolor{black}{#1}}
\newcommand{\ssy}[1]{\textcolor{black}{#1}}

\begin{document}

\title{From Local to Global to Mechanistic: An iERF-Centered Unified Framework for Interpreting Vision Models}
\author{Yearim Kim$^{*}$,
        Sangyu Han$^{*}$,
        and~Nojun Kwak$^{\dagger}$
\thanks{$^{*}$Equal contribution. $^{\dagger}$Corresponding author.}%
\thanks{The authors are with Seoul National University, Seoul 08826, South Korea (e-mail: \{yerim1656, acoexist96, njkwak\}@snu.ac.kr).}}


\markboth{Journal of \LaTeX\ Class Files,~Vol.~14, No.~8, August~2021}%
{Shell \MakeLowercase{\textit{et al.}}: A Sample Article Using IEEEtran.cls for IEEE Journals}


\maketitle

\begin{abstract}
Modern vision models achieve remarkable accuracy, but explaining \emph{where} evidence arises, \emph{what} the model encodes, and \emph{how} internal computations assemble that evidence remains fragmented. We introduce an iERF-centric framework that unifies local, global, and mechanistic interpretability around a single analysis unit: the \textbf{pointwise feature vector (PFV)} paired with its \textbf{instance-specific Effective Receptive Field (iERF)}. On the local side, \emph{Sharing Ratio Decomposition} (SRD) expresses each PFV as a mixture of upstream PFVs via sharing ratios and \emph{propagates} iERFs to construct class-discriminative saliency maps. SRD yields high-resolution, activation-faithful explanations, is robust to targeted manipulation and noise, and remains activation-agnostic across common nonlinearities. For the global view, we introduce \emph{Concept-\yr{Anchored} Feature Explanation} (CAFE), \yr{which utilizes the iERF as a semantic label, grounding abstract latent vectors in verifiable pixel-level evidence.} With CAFE, we address the challenge of non-localized sparse autoencoder latents—especially in Transformers, where early self-attention mixes distant context. To answer \emph{how} representations are composed through depth, we propose the \emph{\yr{Interlayer Concept Graph}} with \emph{Interlayer Concept Attribution} (ICAT), which quantifies concept-to-concept influence while isolating layer pairs; an interlayer insertion/deletion protocol identifies Integrated Gradients as the most faithful instantiation. Empirically, across ResNet50, VGG16, and ViTs, our framework outperforms baselines in \nj{both} fidelity and robustness, successfully interprets dispersed SAE features, and exposes dominant concept routes in correct, misclassified, and adversarial cases. Grounded in iERFs, our approach provides a coherent, evidence-backed map from pixels to concepts to decisions.
\end{abstract}


\section{Introduction}
The rapid advancement of deep learning in image classification has sharpened the tension between state-of-the-art accuracy and the need for transparency.
Rather than treating interpretability as a single problem, we recast it as three complementary questions that require a cohesive answer: \emph{where} evidence comes from (local), \emph{what} the model encodes consistently (global), and \emph{how} internal computations assemble that evidence (mechanistic). 
In this work, we present a unified framework that addresses all three dimensions through a single, consistent lens.

\begin{figure}[ht]
\begin{center}
\includegraphics[width=.9\linewidth]{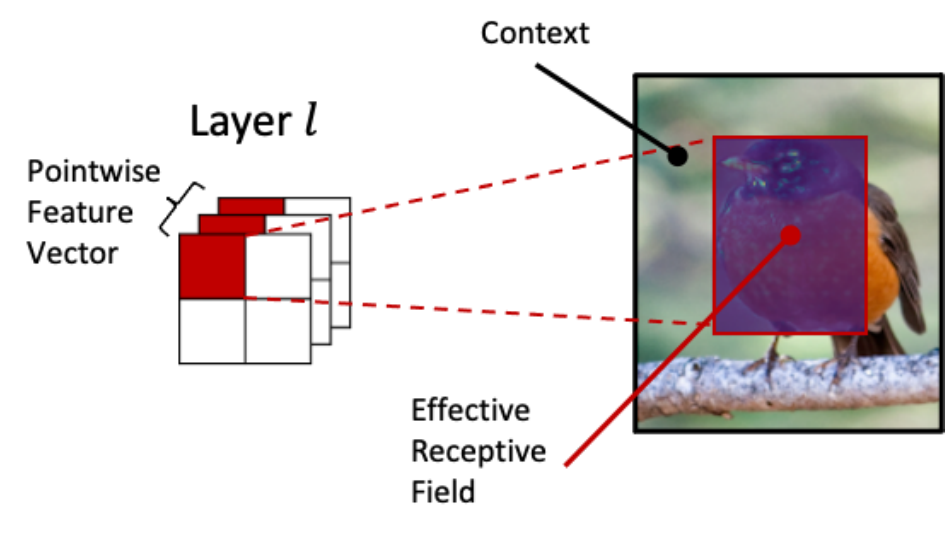}
\end{center}
\vspace{-5mm}
\caption{\yr{\textbf{PFV–iERF bundle} as a spatially-anchored unit for evidence-backed interpretability. \textbf{Pointwise Feature Vector (PFV)}: A PFV is defined as the multi-channel activation vector $v^l_p \in \mathbb{R}^{C^l}$ at a specific spatial coordinate $p$ within hidden layer $l$. \textbf{Instance-specific Effective Receptive Field (iERF)}: For a given input image, the iERF identifies the specific pixel-level attributional evidence that drives the activation of a particular feature.}}
\label{fig:pfv-ierf}
\end{figure}

Our framework is built upon a fundamental analysis unit: the \textbf{Pointwise Feature Vector~(PFV) labeled with its \textbf{instance-specific Effective Receptive Field~(iERF).}}
Following the foundational formulation in our previous work \cite{han2024respect}, a PFV represents the channel-wise activation vector at a specific spatial location, while the iERF maps this vector back to the input pixel space (Fig.~\ref{fig:pfv-ierf}).
\yyr{We characterize these constructs as universal computational units, grounded in the assumption that regardless of the specific architecture—be it the local sliding windows of CNNs or the global self-attention of Vision Transformers (ViTs)—internal representations can be consistently decomposed into spatially-indexed activation vectors. By treating these spatial coordinates as the invariant anchor for attribution, our framework enables a unified analysis of information flow under fixed model parameters.}
On top of this unit, we propose three interconnected methodologies to cover the spectrum of interpretability \emph{from local to global to mechanistic}.

Our framework begins by formalizing local evidence.
We introduce \textbf{Sharing Ratio Decomposition (SRD)~\cite{han2024respect}}, a method that decomposes intermediate features based on their forward contribution to the next layer.
Unlike conventional methods including naive Gradient~\cite{simonyan2014naivegrad}, Grad-CAM~\cite{selvaraju2017gradcam}, \nj{and} LRP~\cite{Montavon2017lrp} that rely on gradients or heuristic post-processing, 
SRD relies solely on model-generated information, yielding robustness to noise and adversarial manipulation while remaining agnostic to the choice of activation function. 

With the fundamental unit established, we extend our scope to the global dataset level to understand what the model encodes.
We first define the \textbf{nodes} of our framework by extracting consistent directions in the PFV space, utilizing methods ranging from clustering to Sparse Autoencoders~(SAE).
However, an extracted concept vector is merely a numerical array lacking inherent semantics.
Thus, to interpret it, we must identify what input patterns trigger it.
While prior works infer meaning through indirect techniques like masking or pooling~\cite{ghorbani2019ace, kowal2024vcc}, we propose using the \textbf{iERF as a direct semantic label} for each vector.
This direct labeling is particularly critical given the challenge of \emph{non-localized} features found in Vision Transformers~(ViTs) or SAE latents~\cite{huben2024sparse}, where activations can be scattered across distant regions due to context mixing.
To resolve this, we introduce \textbf{Concept-Anchored Feature Explanation~(CAFE).}
By utilizing the iERF as a semantic label, CAFE grounds abstract vectors in pixel-level \emph{attributional evidence}, disentangling correlational spatial overlap from true computational reliance.
This bridges the gap between local evidence and global concepts, ensuring faithful interpretation even for dispersed features.

Finally, we bridge these components to trace the mechanistic pathways of the \ssy{model}.
To understand how grounded concepts interact across layers to form high-level decisions, we propose the \textbf{Interlayer Concept Graph}, a directed multigraph whose \emph{nodes} are iERF-anchored concept vectors (PFV–iERF bundles) and whose \emph{edges} quantify \sy{interlayer} influence from parent concepts at layer $a$ to child concepts at layer $b>a$.
We compute these edges via \textbf{Interlayer Concept ATtribution~(ICAT)}, which measures the contribution of a source-layer parent concept to target-layer child concepts.

Using an interlayer insertion/deletion protocol, we identify \emph{Integrated Gradients} as the most faithful instantiation (ICAT\textsubscript{IG}) for quantifying these interactions.

Together, this progression—from where a model looks, to what it encodes, to how those encodings emerge, and finally to why dispersed features activate—forms a continuum of explanation.
Grounded in iERF analysis, our unified framework delivers robust, faithful, and comprehensive interpretability for modern vision models.
Through extensive experiments on CNNs (ResNet50, VGG16) and Transformers (ViT), we demonstrate how these contributions collectively advance our ability to understand and trust deep image classifiers.

\section{Related Work}
\subsection{Local Explanations}
Local explanations answer \emph{where} a model looks at for a given input. 
\yyr{
Gradient-based saliency (naive Gradient~\cite{simonyan2014naivegrad}, Grad$\times$Input~\cite{shrikumar2016not}, IG~\cite{sundararajan2017ig}) compute input-space sensitivity by differentiating the output with respect to input pixels, yielding per-pixel scores without architectural constraints. 
CAM-based methods~(Grad-CAM~\cite{selvaraju2017gradcam}, Grad-CAM++~\cite{chattopadhay2018grad}, Score-CAM~\cite{wang2020score}, Ablation-CAM~\cite{ramaswamy2020ablation}, XGrad-CAM~\cite{ruigang2020xgrad}, Layer-CAM~\cite{jiang2021layercam}) project class-discriminative weights back onto a chosen convolutional feature map, trading spatial resolution for semantic localization. 
Propagation-based attribution approaches~(DeepLIFT~\cite{shrikumar2019deeplift}, LRP and {LRP}$_{z^+}$~\cite{bach2015LRP, Montavon2017lrp}) redistribute the output score layer-by-layer according to predefined propagation rules, aiming to satisfy conservation properties across the network depth.
}

\ssy{Within propagation-based explanations, SRD is closest in spirit to LRP~\cite{bach2015LRP} in its recursive redistribution of evidence. The key difference is the attribution unit: LRP assigns relevance to individual neurons, whereas SRD defines contributions directly between PFVs, the analysis unit used throughout our framework}

\subsection{Global Interpretability}
As the field moved beyond per-instance saliency to ask \emph{what} models encode across datasets, concept-centric approaches emerged. 
Concept-based global interpretability has progressed from directional sensitivity (TCAV)~\cite{kim2018tcav} to automatic concept discovery (ACE/CAT)~\cite{ghorbani2019ace, fel2023craft}, and more recently to sparse autoencoders (SAEs) that learn overcomplete, $\ell_1$-regularized latents adapted to ViTs~\cite{lim2025sparse}.
Yet individual units are often \emph{polysemantic}, and most pipelines still visualize top-activating images/tokens—an activation-centric practice that works only for \emph{localized} features and breaks under the \emph{non-localized} activations induced by attention~\cite{lim2025sparse, zaigrajew2025interpreting, stevens2025scientificallyrigorous}.
We address this by adopting the PFV--iERF analysis unit (treating ViT tokens as PFVs) and anchoring features to their pixel-level provenance~(iERF).
This process moves beyond mere discovery to semantic grounding, resolving the ambiguity of ``non-localized features'' and providing a stable visual basis for abstract units like the Class Token. 
By synthesizing SAE-based discovery with iERF-grounding, we ensure that global explanations reflect the true attributional drivers of the model's internal vocabulary.

\subsection{Mechanistic Interpretability}
Mechanistic interpretability seeks \emph{how} internal computations form task evidence across layers.
A central challenge is defining the computational units, or nodes, that participate in these circuits.
Following the established paradigm of representing concepts as subspaces or vectors in the activation space—as explored by~\cite{Chormai2024pfv1, oma2023disentangling}, we utilize vector decomposition as a fundamental step for node generation.

Our work advances this direction by constructing a unified \textbf{Interlayer Concept Graph (ICG)} with the node.
While CRP~\cite{Achtibat2023CRP} provides high-resolution relevance maps, it remains logit-dependent and often requires predefined concept dictionaries, resulting in fragmented, class-specific views rather than a dataset-wide mechanistic understanding.
Similarly, the Visual Concept Connectome (VCC)~\cite{kowal2024vcc} summarizes inter-concept relations but operates as a global class-wise summary.
Instead, our work enables tracing the instance-specific evolution of evidence through the network. 
By grounding nodes in the PFV--iERF unit, we provide a unified map that resolves conceptual conflicts across both classes and individual instances.

\section{Analytic Foundations}
\label{sec:analysis unit}
In this section, we introduce the two primitives that underpin our framework, the \textbf{Pointwise Feature Vector (PFV)} and the \textbf{Instance-specific Effective Receptive Field (iERF)}, \nj{which were originally presented in our previous work \cite{han2024respect}. Then we} explain why their pairing forms the right analysis unit for local, global, and mechanistic interpretability.

\subsection{Pointwise Feature Vector}
We define a Pointwise Feature Vector (PFV) as the vector of neurons along the channel axis within a hidden layer that share the same receptive field.
Formally, given the embedding of layer $l$, denoted as $A^l \in \mathbb{R}^{H^l W^l \times C^l}$, where $C^l$ is the number of channels and $H^l W^l$ represents the spatial dimensions of the feature map, the PFV at position $p \in \{1, \cdots, H^l W^l\}$ is expressed as $v^l_p \in \mathbb{R}^{C^l}$.

\yr{
We adopt this vector-centric view to address the limitation of individual scalar neurons.
As widely discussed in mechanistic interpretability~\cite{elhage2022superposition, olah2020zoom}, single neurons are often polysemantic, firing for multiple unrelated concepts.
This phenomenon, driven by superposition--where the model packs more features than available dimensions--means that semantic concepts are rarely aligned with standard basis axes.
Instead, they are encoded as distributed representations~(directions) in the high-dimensional activation space~\cite{szegedy2013intriguing, fong2018net2vec}.
Since \sy{a PFV preserves the complete local response of the receptive-field computation as a multi-channel vector, it serves as the natural unit for analyzing concept-related signals at the vector level rather than as individual neuron readings.}}
\ssy{This unit choice also matters methodologically: when the explanatory target is a PFV rather than an individual neuron, neuron-wise relevance does not directly yield a canonical PFV-level score. For detail, refer to Appendix ~\ref{app:pfv_neuronwise_note}.}

\yr{While the utility of hidden representations for interpretability is well-recognized, existing methods often sacrifice spatial precision for mathematical abstraction.
For instance, \cite{Kauffmann2024pfv2} and \cite{Chormai2024pfv1} utilize hidden vectors for cluster explanation and subspace disentanglement, yet they do not explicitly preserve the spatial correspondence of these features.
Similarly, automated concept discoverers like ACE~\cite{ghorbani2019ace} often rely on spatial pooling or superpixel clustering, which can obscure the explicit $(x, y)$ coordinates and the precise input-level evidence.
TCAV~\cite{kim2018tcav} defines concepts as vectors in the activation space but requires user-defined datasets and yields a global score, discarding local spatial information.}

\yr{In contrast, our PFV formulation strictly maintains spatial locality.
By anchoring each feature vector to its specific coordinates,} we preserve the ``where" information that is critical for vision tasks.
Furthermore, this abstraction provides a unified framework for both CNNs and Vision Transformers (ViTs); in ViTs, the token embedding naturally serves as a PFV, allowing our framework to apply seamlessly across architectures without the loss of spatial context inherent in global subspace projections.

\subsection{Instance-specific Effective Receptive Field (iERF)}
While theoretical receptive fields specify the maximum input region that could influence a unit, it often expands to cover—or even exceed—the entire image, making it too coarse for explanation \cite{araujo2019computingrf}.
Empirical analysis further shows that the actual influence concentrates within a much smaller, approximately Gaussian region—termed the \emph{Effective Receptive Field (ERF)}~\cite{luo2016erf}.
However, these notions still do not identify, for a given instance, which \yr{specific} pixels \yr{drive} a particular feature to fire—hence the need for instance-specific, evidence-anchored attributions.

We define the instance-specific ERF~(iERF) of a feature $k$ in image $I$ as a score map capturing the \emph{attributional evidence} for that feature:
\begin{equation}
    \text{iERF}_k(I) = \{(p, A(p \mid z_k, I)) : p \in I\},
\end{equation}
where $A(p \mid z_k, I)$ denotes the attribution of input \nj{pixel (or patch)} $p$ to the scalar output $z_k(I)$.
\yr{This formulation distinguishes our approach from cluster-based explanations~\cite{kau2022clustering}.
While such methods interpret features by mapping them to \textbf{fixed, pre-computed prototypes} (static), our iERF traces the \textbf{actual computational path} unique to each specific input (dynamic).}
This distinction is crucial for identifying which pixels \emph{actually} drive a feature's activation, \sy{enabling evidence-backed interpretation of the vector by identifying which input regions support its activation.} 

\subsection{PFV--iERF as the Unified Analysis Unit}
We now motivate why pairing a Pointwise Feature Vector (PFV) with its instance-specific Effective Receptive Field (iERF) is the right primitive for our framework.
A PFV encodes the \emph{what} (the numerical feature state), but without context, it remains an abstract array of numbers.
Its iERF supplies the \emph{label} (the pixel-level provenance), explicitly assigning semantic meaning to the vector.
Together, they form a single analysis unit that spans the three questions—\emph{where}, \emph{what}, and \emph{how}—underlying local, global, and mechanistic explanations.

\paragraph{Local perspective (Where)}
Pairing PFVs with iERFs yields fine-grained, activation-faithful saliency.
Because a PFV retains the joint multi-channel state at one location, its iERF identifies the precise pixels that \nj{are} responsible for that state, producing faithful maps without relying on gradient heuristics or post-hoc smoothing.

\paragraph{Global perspective (What)}
Within a fixed layer $l$, the PFV space $\mathcal{V}(l)$ provides a layer-wise geometry for discovering global concepts~(nodes). 
For each layer $l$, we define a concept as a linear basis in the high-dimensional PFV space \sy{$\mathcal{V}(l)$} that represents semantically meaningful features across different images, transcending classes.
Thus, aggregating PFVs across images and clustering in PFV space yields layerwise concept vectors.
However, interpreting these nodes is non-trivial.
Distinct from methods that infer meaning indirectly via bounding boxes~\cite{fel2023craft} or segmentation proxies~\cite{ghorbani2019ace, fel2024holistic}, we employ the iERF as a \textbf{direct semantic label}.

Furthermore, unlike prior global approaches that tacitly equate a concept with the top-activated patch/token or its immediate visual neighborhood~\cite{zaigrajew2025interpreting, pach2025saevlm, stevens2025scientificallyrigorous}, the iERF  identifies the input region driving the activation regardless of the token's spatial position.
Specifically, in ViTs where tokens mix global context, the iERF disentangles \emph{which pixels} truly drive a token's state, enabling faithful semantic labeling even when the evidence is non-local.

\paragraph{Mechanistic perspective (How)}
The PFV--iERF unit serves as the node for interlayer analysis. 
By tracing how labeled PFV-based concepts decompose and recombine across layers, we map the mechanistic pathways of decision-making.
Thus, the PFV--iERF unit acts as the connective tissue linking local evidence, global structure, and mechanistic interactions, offering a more rigorous and unified currency than disparate methods used in isolation.

\subsection{Implementation Details}
\paragraph{Extending to Vision Transformers}
\yr{\nj{The PFV-iERF framework originally developed for CNNs in \cite{han2024respect} can be naturally extended} to Vision Transformer~(ViTs).}
Because a convolutional layer can be viewed as a specific form of multi-head attention layer~\cite{Cordonnier2020On}, every token vector $t^l_q$ can likewise be regarded as a PFV.
\yr{Given the context-mixing nature of self-attention, iERFs are essential for identifying the true source of a token's activation.
For attribution, we adopt \textbf{AttnLRP}~\cite{achtibat2024attnlrp} for ViTs, as it provides faithful token-level contributions, while using SRD \cite{han2024respect} for CNNs.} \ssy{For ViTs, AttnLRP provides channel-wise token attributions; when a scalar token-level score is required to instantiate an iERF, we summarize these attributions by an $l_2$
-norm across channels}


\paragraph{\yr{Relevance-Weighted Sampling}}
To construct a meaningful PFV--iERF dataset, we must avoid overrepresenting uninformative background features (e.g., sky, grass).
Instead of uniform sampling, we sample a single PFV from each image in proportion to its contribution to the final output logits, yielding 50,000 PFV–iERF pairs per layer.
This \emph{relevance-weighted sampling} ensures that our analysis focuses on semantically meaningful features that actively drive the model's predictions, thereby strengthening the fidelity of both extracted concepts and mechanistic graphs.

\section{Local Explanation}
\label{local}

To establish a unified interpretability framework, we present a methodology aimed at quantifying local, pixel-level evidence by focusing on the instance-specific Effective Receptive Field (iERF).
While theoretical receptive fields provide a maximum bound of influence, the iERF seeks to identify the specific input features that drive a particular activation for a given instance.
We utilize Sharing Ratio Decomposition (SRD) as an analytic engine to propagate and aggregate these fields across the network layers.

\subsection{Saliency Map Synthesis via iERF Aggregation}
\begin{figure*}[ht]
\begin{center}
\includegraphics[width=0.9\linewidth]{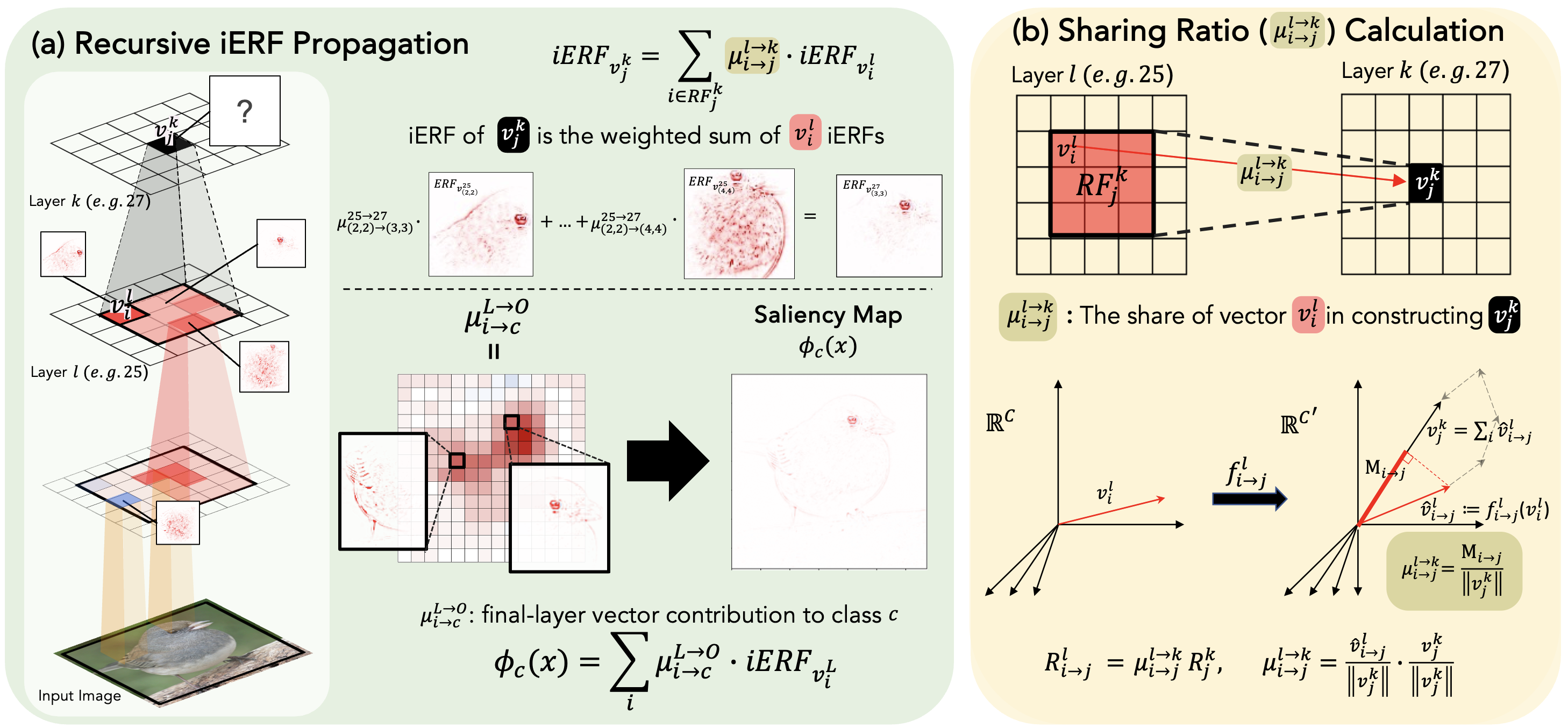}
\end{center}
\caption{\yr{\textbf{Local explanation map synthesis through Recursive iERF Propagation and Sharing Ratio Decomposition.} 
\textbf{(a) Recursive iERF Propagation}: Schematic of the layer-wise construction of the instance-specific Effective Receptive Field~(iERF). The iERF of Pointwise Feature Vector~(PFV) $v^{k}_j$ at layer $k$ is computed as \ssy{the} weighted sum of iERFs from PFVs ($v^{i}_l$) in the preceding layer $l$ that fall within its receptive field $RF_j^k$. This process is applied recursively from the input image to the final encoder layer $L$, where the class-specific saliency map $\phi_c(x)$ is synthesized by aggregating the iERFs based on their contribution $\mu^{L \rightarrow O}_{i \rightarrow c}$ to class $c$.
\textbf{(b) Sharing Ratio ($\mu^{l \rightarrow k}_{i \rightarrow j}$) Calculation}: To quantify the contribution of a source PFV $v^l_i$ to a target PFV $v^k_j$, we decompose $v^k_j$ into partial components $\hat{v}^l_{i \rightarrow j}$ using the model's internal affine transformations ($f^l_{i \rightarrow j}$). The sharing ratio is then defined as the normalized inner product (directional alignment) between the partial contribution $\hat{v}^l_{i \rightarrow j}$ and the target vector $v^k_j$.}}
\label{fig:method_fp}
\end{figure*}

\yr{As shown in \ssy{Fig.~\ref{fig:method_fp}}, the synthesis of local explanations begins with the recursive construction of iERFs, which represent the input-level attribution evidence forming the current state Pointwise Feature Vector~(PFV).
A PFV represents the multi-channel activation state at a specific spatial location, and its corresponding iERF maps this state back to the input space.
In a feed-forward network, the activation of a PFV $v^k_j$ at layer $k$ is determined by the PFVs in the preceding layer $l$ that fall within its receptive field, denoted as $RF^k_j$.
Consequently, we sequentially build the iERF of a PFV $v^k_j$ by decomposing it into a weighted combination of iERFs from the preceding layer's spatial grid:
\begin{equation}
    iERF_{v^k_j} = \sum_{i \in RF_j^k} \mu^{l \rightarrow k}_{i \rightarrow j} \cdot iERF_{v^l_i},
\end{equation}
}
\noindent where the base case is defined as $iERF_{v^0_i} = E^i$~(the unit matrix \nj{whose $i$-th element is 1 and all the others are 0's.}), signifying that the attributional basis for an input pixel is the pixel itself.
This recursive process is intended to ensure that the resulting saliency map for class $c$, $\phi_c(x)$ captures the precise local contributions of each input pixel to the final decision.

The operationalization of this construction requires the estimation of the \textbf{sharing ratio} $\mu^{l \rightarrow k}_{i \rightarrow j}$, which quantifies the relative contribution of a source PFV $v^l_i$ to a target PFV $v^k_j$.
By utilizing the model's internal affine transformations, we decompose the target PFV into partial contributions $\hat{v}^l_{i \rightarrow j} = f^l_{i \rightarrow j}(v^l_i)$.
The sharing ratio is then calculated as the normalized inner product between this partial contribution and the target PFV:
\begin{equation}
    \mu_{i\rightarrow j}^{l\rightarrow k} = \Big\langle \frac{\hat{v}_{i \rightarrow j}^{l}}{\lVert v_j^{k} \rVert}, \frac{v_j^{k}}{\lVert v_j^{k} \rVert} \Big\rangle.
\end{equation}
This formulation captures the directional alignment between source and target features while ensuring that $\sum_{i \in RF^k_j}\mu^{l \rightarrow k}_{i \rightarrow j}=1$, thereby providing a conservative redistribution of attributional weight.

Consequently, the iERFs can be recursively constructed layer by layer until reaching the encoder output.
At this \nj{encoder output (layer $L$)}, the representation consists of \nj{$H^L W^L$} PFVs, each associated with its corresponding iERF.
The final class-specific saliency map $\phi_c(x)$ is then obtained as a weighted sum of the iERFs at the encoder output:
\begin{equation}
\phi_c(x) = \sum_{i} \mu^{L \rightarrow O}_{i \rightarrow c} \cdot iERF_{v^L_i},
\end{equation}
where $\mu^{L \rightarrow O}_{i \rightarrow c}$ denotes the contribution of PFV $v^L_i$ at the final encoder layer $L$ to the output logit of class $c$.
Since the subsequent MLP head flattens vectors into scalars, our vector-based formulation is no longer necessary at this stage.
For the classifier, we therefore adopt established gradient-based attribution methods such as Grad-CAM~\cite{selvaraju2017gradcam}.

\begin{figure*}[t]
\begin{center}
\includegraphics[width=.9\linewidth]{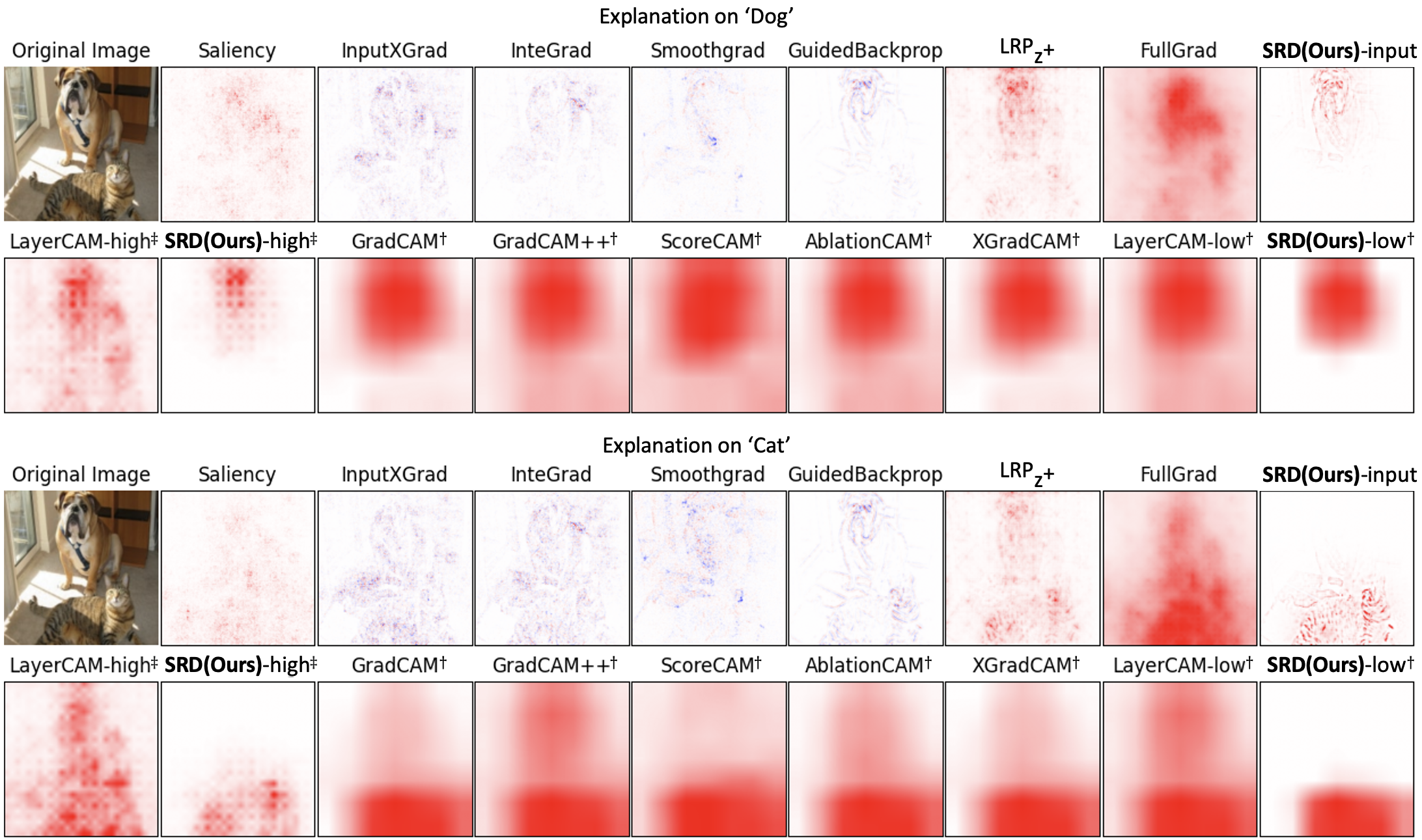}
\end{center}
\vspace{-5mm}
\caption{\textbf{Qualitative comparisons on ResNet50.} \textbf{Top:} class “dog”; \textbf{bottom:} class “cat.” Methods marked with † operate at $7{\times}7$ resolution; those marked with ‡ at $28{\times}28$; all others are at input scale ($224{\times}224$). SRD (input-scale) preserves fine image details better than competing methods.}
\label{fig:qual}
\end{figure*}
\sy{To obtain class-discriminative final sharing ratios, we first compute a PFV-level class contribution score $\Phi_i^c$ for each PFV $v_i^L$ using a Grad-CAM--style channel-weighted sum (Eq.~(6)).
We then convert these scores into the final sharing ratios $\mu^{L \rightarrow O}_{i \rightarrow c}$ used in Eq.~(4) by centering $\Phi_i^c$ with respect to the mean score across classes at the same PFV and taking the positive part (Eq.~(\ref{eq:mu_last})).
\footnote{The sum of modified sharing ratios is not guaranteed to equal one.}
Intuitively, the mean-subtraction suppresses PFVs that contribute similarly across many classes (class-agnostic signals), while emphasizing PFVs that are relatively more supportive of class $c$.
The $\max(\cdot,0)$ operator yields a \emph{positive-evidence} saliency map.
\sy{We analyze the effect of this refinement at Appendix~\ref{appendix:mu_ablation}.}
\begin{equation}
\mu^{L \rightarrow O}_{i \rightarrow c}
= \max\!\left(\Phi_i^c - \frac{1}{K}\sum_{c' \in [K]} \Phi_i^{c'},\, 0\right),
\label{eq:mu_last}
\end{equation}
with
\begin{equation}
\Phi_i^c = \sum_{k} \alpha^c_k A^k_i, \qquad
\alpha^c_k = \frac{1}{H^LW^L}\sum_{i \in [H^LW^L]} \frac{\partial y^c}{\partial A^k_i}.
\end{equation}
Here, $A^k_i$ is the $k$-th element of PFV $v^{L}_i$, and $y^c$ is the $c$-th element of the output logit vector $y \in \mathbb{R}^K$.}

\subsection{Experiments}

We compared our proposed Sharing Ratio Decomposition (SRD)~\cite{han2024respect} against a broad spectrum of attribution and CAM-based methods, including naive Gradient~\cite{simonyan2014naivegrad}, Guided Backpropagation~\cite{springenberg2014striving}, Integrated Gradients~\cite{sundararajan2017ig}, LRP$_{z^+}$~\cite{Montavon2017lrp}, SmoothGrad~\cite{daniel2017smoothgrad}, FullGrad~\cite{srinivas2019full}, GradCAM~\cite{selvaraju2017gradcam}, GradCAM++~\cite{chattopadhay2018grad}, ScoreCAM~\cite{wang2020score}, AblationCAM~\cite{ramaswamy2020ablation}, XGradCAM~\cite{ruigang2020xgrad}, and LayerCAM~\cite{jiang2021layercam}. 
These baselines were carefully selected to represent the major families of explanation methods ensuring that our evaluation addresses the landscape of state-of-the-art local interpretability techniques. 

\paragraph{Experimental setup}  
We employed ResNet50 and VGG16 as representative CNN architectures. 
To account for the inherent differences in map resolution across methods, we followed the best-practice layer choices suggested by prior work, producing three saliency map scales: low-resolution $(7 \times 7)$, high-resolution $(28 \times 28)$, and input-scale $(224 \times 224)$. 
All saliency maps were normalized and upsampled to the input scale for fair comparison.

\subsubsection{Qualitative Results}
Fig.~\ref{fig:qual} illustrates representative explanations for images containing two classes, a cat and a dog. 
We observe that conventional methods often generate highly similar maps across classes, failing to isolate class-specific evidence. 
In contrast, SRD produces counterfactual and fine-grained explanations, sharply delineating the features that distinguish ``dog" from ``cat". 
This qualitative advantage is consistent across diverse samples (see the supplementary materials), and highlights SRD’s capacity to go beyond generic heatmaps to yield class-discriminative insights.

\subsubsection{Quantitative Results}
\label{subsubsec: local_quan}
For a systematic comparison, we evaluated all methods on the ImageNet-S50 dataset~\cite{gao2022large}, which contains 752 samples with their object segmentation masks. 
We assessed four complementary dimensions: (i) \emph{localization} (Pointing Game~\cite{zhang2018top}, Attribution Localization~\cite{kohlbrenner2020towards}), (ii) \emph{complexity} (Sparseness~\cite{chalasani2020concise}), (iii) \emph{faithfulness} (Fidelity~\cite{ijcai2020p417}), and (iv) \emph{robustness} (Stability~\cite{alvarez2018towards}). 
Metric definitions are provided in the supplementary materials. 

As summarized in Tab.~\ref{tab:quantitative}, SRD consistently outperformed competing methods.  
On VGG16, SRD-high achieved the best scores in Pointing Game and Fidelity, while SRD-input excelled in Sparseness and Stability, reflecting both functional precision and resilience.  
On ResNet50, where residual connections notoriously degrade CAM-based approaches, many baselines suffered substantial performance drops despite the stronger backbone.  
Remarkably, SRD remained competitive: SRD-high led in Attribution Localization and Fidelity, while SRD-input ranked best in Pointing Game and Stability.  
This indicates that SRD’s decomposition is robust not only to architectural variations but also to structural challenges such as skip connections.  
\begin{table*}[t]
    \centering
    \caption{\textbf{Average results on ImageNet-S50 ($n{=}752$ images) for five metrics}: Pointing Game (Poi.), Attribution Localization (Att.), Sparseness (Spa.), Fidelity (Fid.), and Stability (Sta.). All metrics are computed on normalized saliency maps following the default settings of \cite{hedstrom2023quantus}. Methods marked with † operate at $7{\times}7$ resolution; those marked with ‡ at $28{\times}28$; all others use input scale ($224{\times}224$). Best scores are in \textbf{bold}; second-best are \underline{underlined}.} 
    \scalebox{0.83}{
        \begin{tabular}{l c c ccccc c ccccc}
        \toprule
        &&& \multicolumn{5}{c}{VGG16} &&  \multicolumn{5}{c}{ResNet50}  \\
        \cmidrule(lr){4-8} \cmidrule(lr){10-14} 
        
        Method& && Poi.$\uparrow$ & Att.$\uparrow$ & Spa.$\uparrow$ & Fid.$\uparrow$ & Sta.$\downarrow$
        && Poi.$\uparrow$ & Att.$\uparrow$ & Spa.$\uparrow$ & Fid.$\uparrow$ & Sta.$\downarrow$ \\
        \midrule
        Saliency &   && .793 & .394 & .494 & .093 & .181
                  && .654 & .370 & .488 & .063 & .172 \\
        GuidedBackprop &   && .892 & .480 & \underline{.711} & .022 & \underline{.100}
                  && .871 & .498 & \textbf{.741} & .022 & \underline{.112} \\
        GradInput &   && .781 & .387 & .630 & -.013 & .181 
                  && .639 & .361 & .626 & -.018 & .178 \\
        InteGrad &   && .869 & .416 & .618 & -.017 & .175 
                  && .759 & .382 & .614 & -.016 & .171 \\
        $\textrm{LRP}_{z^+}$  &   && .855 & .456 & .535 & .098 & .182 
                  && .543 & .332 & .572 & .012 & .105 \\
        Smoothgrad &   && .845 & .363 & .536 & -.005 & .190
                  && .888 & .396 & .556 & -.002 & .166 \\
        Fullgrad  &   && .796 & .362 & .334 & .107 & .203
                  && .938 & .387 & .262 & .123 & .689 \\
        $\textrm{GradCAM}^{\dagger}$ &   && \underline{.945} & .431 & .466 & .175 & .583
                  && .946 & .424 & .411 & .128 & .757 \\
        $\textrm{GradCAM++}^{\dagger}$ &   && .932 & .429 & .351 & .176 & .570
                  && .945 & .414 & .386 & .129 & .732 \\
        $\textrm{ScoreCAM}^{\dagger}$ &   && .937 & \textbf{.582} & .342 & .167 & .622
                  && .916 & .381 & .313 & .123 & .827 \\
        $\textrm{AblationCAM}^{\dagger}$ &   && .928 & .481 & .493 & .189 & .622
                  && .934 & .394 & .329 & .133 & .814 \\
        $\textrm{XGradCAM}^{\dagger}$ &   && .896 & .406 & .446 & .181 & .576
                  && .946 & .424 & .411 & .126 & .753 \\
        $\textrm{LayerCAM-low}^{\dagger}$ &   && .869 & .425 & .446 & .175 & .450
                  && .934 & .411 & .379 & .128 & .734 \\
        $\textrm{LayerCAM-high}^{\ddagger}$ &   && .865 & .435 & .401 & \underline{.199} & .423
                  && .941 & .423 & .349 & \underline{.135} & .486 \\
        \midrule
        
        $\textbf{SRD-low}~(ours)^{\dagger}$     & && \underline{.945} & .424 & .437 & .179 & .595
                  && .946 & .544 & .682 &  .130 &  .600 \\
        $\textbf{SRD-high}~(ours)^{\ddagger}$     & && \textbf{.948} & \underline{.566}& .629 & \textbf{.206} & .406
                  && \underline{.952} & \textbf{.579} & .628 &  \textbf{.142} &  .375 \\ 
        \textbf{SRD-input}~(ours)     & && .925 & .561 & \textbf{.788} & .069 & \textbf{.099}
                  && \textbf{.953} & \underline{.576} & \underline{.724} &  .082 &  \textbf{.104} \\    
        \bottomrule 
        \end{tabular}
    }
    \label{tab:quantitative}
    \vspace{-3mm}
\end{table*}

\subsubsection{Adversarial Robustness}
\begin{figure}[ht]
\begin{center}
\includegraphics[width=.95\linewidth]{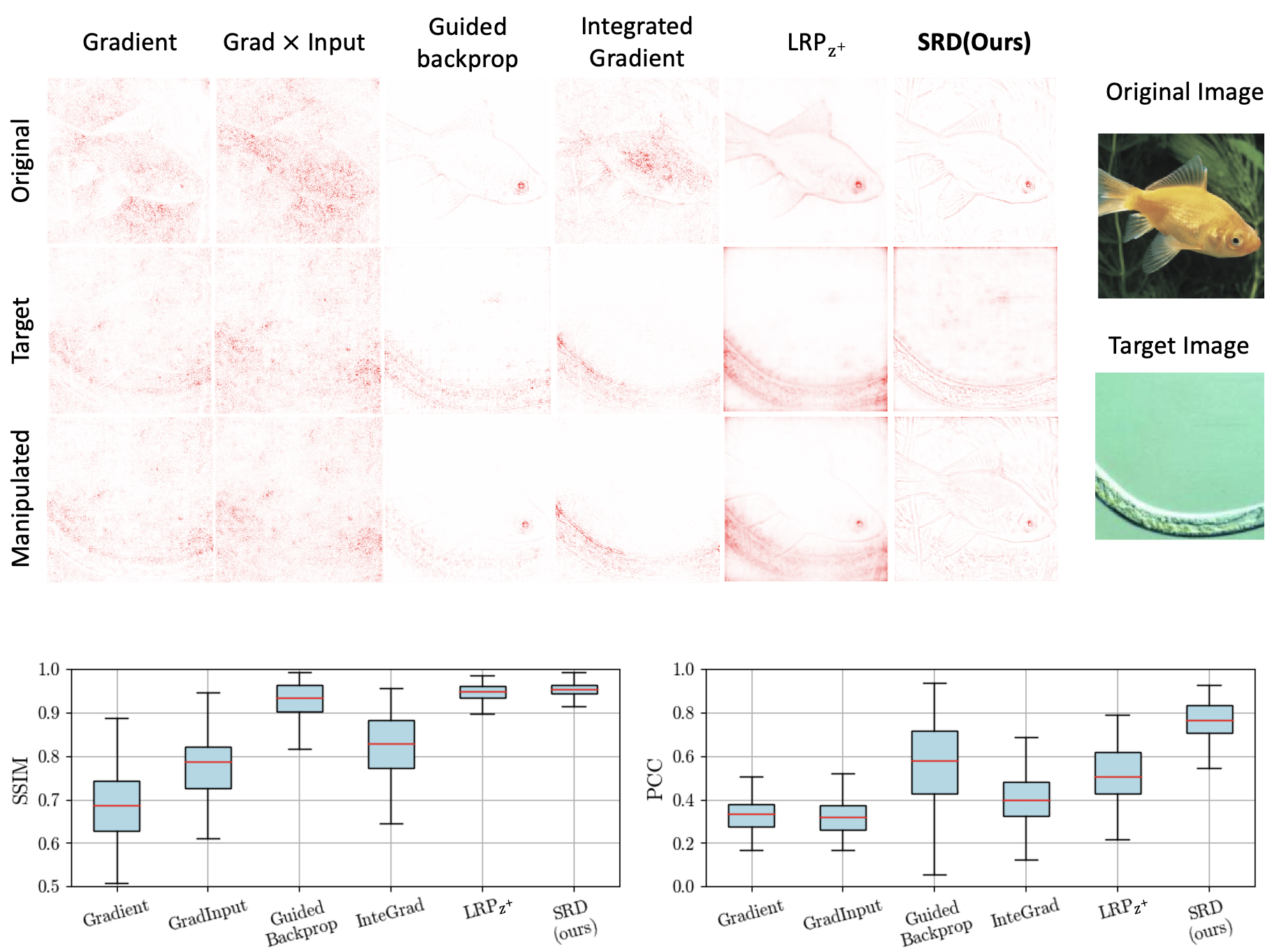}
\end{center}
\vspace{-5mm}
\caption{\textbf{Adversarial manipulation of explanations.} \textbf{Top:} Qualitative comparison under a targeted attack that preserves model logits; competing methods remove class-relevant evidence (goldfish), whereas SRD preserves it (additional examples in the supplementary materials). 
\textbf{Bottom:} Quantitative robustness—higher Structural Similarity (SSIM) and Pearson correlation (PCC) indicate lower susceptibility to manipulation; SRD attains the best scores on both. Best viewed when enlarged.}

\label{fig:exp_robust}
\end{figure}
Explanations can be adversarially manipulated by imperceptible input perturbations while keeping the model's prediction essentially unchanged~\cite{dombrowski2019explanations}, casting doubt on their reliability.
We therefore evaluate robustness against \emph{targeted} explanation attacks (Fig.~\ref{fig:exp_robust}). 

Following \cite{dombrowski2019explanations}, the perturbation $\delta$ is optimized to minimize
\begin{equation}
    \mathcal{L} = \lambda_1 \left\| \phi(x_{adv})-\phi(x_{target})\right\|^2 + \lambda_2 \, \left\| F(x_{adv})-F(x_{org})\right\|^2,
\end{equation}
where $x_{\mathrm{adv}}=x_{\mathrm{org}}+\delta$, $\phi(x)$ denotes the saliency map for image $x$, and $F(x)$ is the model’s logit output.
We use $\lambda_1=1e11$ and $\lambda_2=1e6$ as in the original setup.

We conducted targeted manipulation on 100 randomly selected ImageNet image pairs with VGG16.
Because the attack requires gradient-trackable explanations, we compare \nj{our \emph{SRD}} against \emph{Gradient}, \emph{Gradient$\times$Input}, \emph{Guided Backpropagation}, \emph{Integrated Gradients} \nj{and} \emph{LRP}${z^{+}}$. 
The learning rate is $10^{-3}$ for all methods.
Optimization stops when the MSE between $x$ and $x_{\mathrm{adv}}$ reaches $0.001$; additionally, the per-channel RGB change is bounded by $8$ (on the 0–255 scale) so that $x_{\mathrm{adv}}$ remains visually indistinguishable from $x$.
We take absolute saliency values, as in \cite{dombrowski2019explanations}. To avoid confounding—since $\mu^{L\rightarrow O}_{i\rightarrow c}$ can be estimated using other attributors—we set all $\mu^{L\rightarrow O}_{i\rightarrow c}=1$ during evaluation.
 
Robustness is quantified by the similarity between the original map $\phi(x_{\mathrm{org}})$ and the manipulated map $\phi(x_{\mathrm{adv}})$ using Structural Similarity (SSIM) and Pearson Correlation Coefficient (PCC).
Higher values indicate that the manipulated map preserved the original evidence, thereby demonstrating the robustness of the explanation method.

As shown in \ssy{Fig.}~\ref{fig:exp_robust}, SRD achieves the highest SSIM and PCC among input-scale methods and preserves class-relevant cues under attack (goldfish) that other methods discard (Fig.\ref{fig:exp_robust}).
Combined with our \emph{Stability} results under random noise (Tab.~\ref{tab:quantitative}), these findings indicate that SRD provides not only faithful but also manipulation-resistant explanations.

\subsubsection{Application to various activations}
\label{extension}

\begin{table}[tb]
\centering
\setlength{\tabcolsep}{3pt}         
\renewcommand{\arraystretch}{0.95}
\caption{\textbf{Fidelity across activation functions} on CIFAR-100 with ResNet50. SRD achieves the best fidelity for all tested activations. Corresponding model accuracies are: 0.780 for ReLU, 0.746 for ELU, 0.785 for LeakyReLU, 0.756 for Swish, 0.767 for GeLU, and 0.685 for Tanh.}
\label{tab:activation}
\begin{tabular}{lcccccc}
\hline
Activation                    & ReLU           & ELU            & LeakyReLU      & Swish          & GeLU           & Tanh           \\
\hline
GuidedBackprop                & \underline{0.064}    & 0.025          & 0.001          & 0.015          & 0.030          & 0.028          \\
GradInput                     & -0.010         & -0.007         & -0.005         & -0.024         & -0.004         & -0.006         \\
InteGrad                      & 0.006          & 0.015          & -0.001         & -0.008         & -0.007         & 0.014          \\
LRP$_{z^+}$ & 0.039          & -              & -              & -              & -              & -              \\
Smoothgrad                    & -0.012         & 0.026          & -0.014         & -0.023         & -0.009         & -0.017         \\
Fullgrad                      & 0.038          & \underline{0.209}    & 0.029          & \underline{0.171}    & \underline{0.095}    & \underline{0.107}    \\
GradCAM                       & 0.005          & -0.014         & -0.004         & 0.042          & -0.001         & 0.002          \\
ScoreCAM                      & 0.013          & 0.052          & \underline{0.031}    & 0.061          & 0.010          & 0.017          \\
AblationCAM                   & 0.020          & 0.024          & 0.003          & 0.015          & 0.033          & 0.012          \\
XGradCAM                      & 0.007          & 0.011          & 0.018          & 0.028          & 0.012          & 0.017          \\
LayerCAM                      & 0.021          & 0.042          & 0.012          & 0.018          & 0.007          & -0.001         \\ \hline
SRD(Ours)                     & \textbf{0.078} & \textbf{0.214} & \textbf{0.065} & \textbf{0.194} & \textbf{0.128} & \textbf{0.115} \\
\hline
\end{tabular}
\end{table}

Most prior attribution methods are tailored to ReLU and often require redesign to handle other nonlinearities. 
However, as in Tab.~\ref{tab:activation}, SRD can be applied to various activations due to the utilization of preactivation, while maintaining high fidelity. 
\yr{Additional qualitative results are provided in Appendix~\ref{appendix: qual_activation}.}

\section{Global Explanation: Concept-Anchored Feature Explanation}
\label{sec:global_erf}
While local explanations answer \emph{where} a model looks for a single instance, global interpretability instead asks \emph{what} a model encodes \emph{consistently} across the dataset. 
\yr{However, a fundamental bottleneck in global discovery is the reliance on correlation: interpreting a feature by inspecting top-activating patches.}
We posit that the appropriate analysis unit for this global view is the \textbf{pointwise feature vector~(PFV) labeled with its instance-specific ERF (iERF)}.
In convolutional architectures, spatial locality biases each layer to aggregate information within gradually expanding receptive fields. 
Transformers, by contrast, \emph{mix global information early} through self-attention.
Consequently, the physical location of a token may not align with its semantic driver.
By anchoring abstract feature vectors with iERFs, we can disentangle which pixels actually drive a token’s state, yielding faithful global concepts even when evidence is spatially dispersed.

\subsection{The Challenge of Non-localized SAE Features}
\label{sec:cafe}

To generalize instance-specific attributions into a reusable dataset-level vocabulary, the high-dimensional PFV space must be factorized into discrete, interpretable units.
In the context of Transformer encoders, each layer’s token embedding—including both patch and class tokens—functions as a PFV, representing a multi-channel activation state grounded in a specific receptive field.
For readability, we henceforth use the terms ``token" and ``PFV" interchangeably when discussing Transformer architectures.

Sparse autoencoders (SAEs) \nj{\cite{pach2025saevlm}} emerged as the standard tool for this factorization: overcomplete, sparsity-regularized codes of hidden states surface latents that recur across images and correlate with parts, textures, and attributes.
\yr{For a given hidden representation from a backbone $\textbf{h}\in\mathbb{R}^{d}$, the SAE computes latent activations as:
\begin{equation}
\mathbf{z}=\textrm{ReLU}\big(W_e(\mathbf{h}-\mathbf{b}_d)\big),
\qquad
\hat{\mathbf{h}}=W_d\,\mathbf{z}-\mathbf{b}_h,
\end{equation}
where $W_e\!\in\!\mathbb{R}^{m\times d}$, $W_d\!\in\!\mathbb{R}^{d\times m}$ \nj{for $m \gg d$}, and $\mathbf{b}_d,\mathbf{b}_h$ are learned biases. 
We minimize
\begin{equation}
\mathcal{L}
= \big\lVert \mathbf{h}-\hat{\mathbf{h}}\big\rVert_2^{2}
+ \lambda\lVert \mathbf{z}\rVert_{1},
\end{equation}
so that each latent $z_k$ activates sparsely for a specific visual regularity. 
}
\begin{figure}[t]
    \centering
    \includegraphics[width=0.9\linewidth]{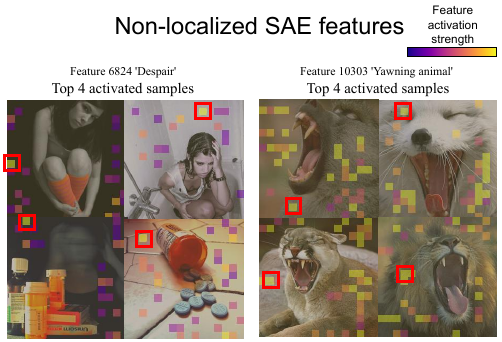}
    \caption{\sy{\textbf{Non-localized SAE feature examples}. For each latent, we show the four most activating samples with token-level activation strength overlaid. Red boxes mark the highest-activation token in each image. In both Feature 6824 (“Despair”) and Feature 10303 (“Yawning animal”), activation is spatially dispersed across multiple distant regions.}} 
    \label{fig:nonlocal_features}
\end{figure}
However, SAE-based discovery exposes a recurring pitfall—some latents are \emph{non-localized}, activating across distant regions and confounding meaning.
\yr{Traditional inspection methods typically rank image patches by their activation magnitudes $z_k(I)$ \sy{and visually \ssy{check} where the activation map is strong. This works reasonably well when a latent is spatially concentrated. But when a latent is non-localized, the visualization supports only a correlational reading—\textit{this feature tends to appear in images like these}—without reliably indicating what input evidence actually drives the latent.} 
\ssy{Fig.}\ref{fig:nonlocal_features} illustrates this failure mode for two non-localized SAE latents. Across their top-activated samples, activation is distributed across semantically unrelated regions, and the single most activated token (red box) can land on an incidental location that is hard to justify as the feature’s semantic cause. }
This spatial dispersion can confound the interpretation of a feature's meaning if one relies solely on top-activated tokens.

Also, this challenge is most pronounced when analyzing features associated with the \textbf{class token ([CLS])}.
Unlike patch tokens, which possess an initial spatial correspondence, the class token is a learned summary vector designed to aggregate global information through self-attention.
Consequently, the class token is inherently 
\nj{not tied to any spatial location;} 
it lacks a fixed origin in the input image plane.
This makes the traditional ``top-activated patch" approach logically inapplicable to class token features, as there is no representative patch to display.
Without a pixel-level trace like the iERF, the visual basis of these abstract, globally-mixed latents remains speculative.

To address this, we instantiate the token–iERF pairing in SAE latents and introduce \textbf{Concept-Anchored Feature Explanation (CAFE)}, \yr{a framework that grounds the dataset-wide latents of Sparse Autoencoders (SAEs) into the pixel-level evidence provided by iERFs.}
\yr{This synthesis addresses a fundamental bottleneck in global interpretability: while SAEs excel at discovering a dictionary of concepts, they often lack a verifiable spatial link to the input, leading to groundless or ambiguous interpretations—especially in globally-mixed architectures like Transformers.}
\yr{By applying iERF-based grounding to SAE latents, CAFE shifts the interpretability paradigm from mere activation-ranking to \textit{attributional provenance}. This ensures that every discovered global concept is not just an abstract direction in a latent space, but a stable semantic anchor grounded in verifiable input-level evidence. This grounding is what allows CAFE to resolve non-localized phenomena and interpret spatially dimensionless units like the Class Token, providing a level of faithfulness that independent global or local methods cannot achieve in isolation.}
\subsection{Concept-Anchored Feature Explanation~(CAFE)}
\begin{figure*}[ht]
    \centering
    \includegraphics[width=0.9\textwidth]{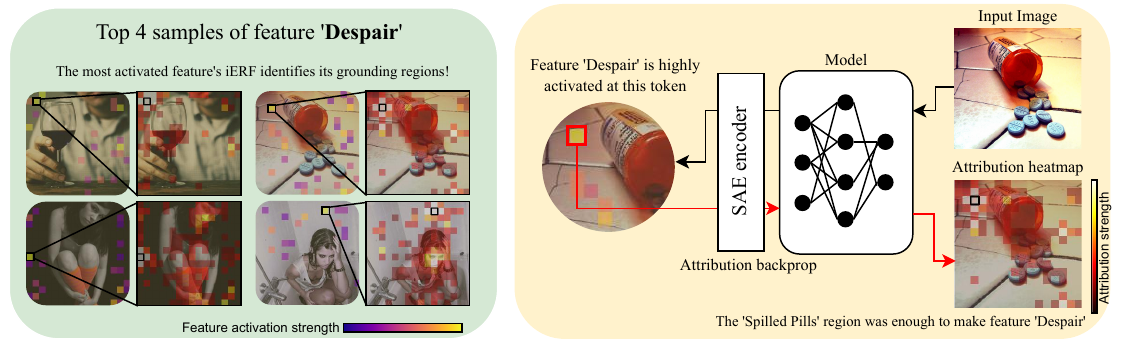}
    \caption{\textbf{Overview of the Concept-Anchored Feature Explanation (CAFE) pipeline.} CAFE identifies the visual provenance of abstract SAE features by computing their instance-specific Effective Receptive Fields (iERFs). For the feature \emph{Despair}, while the maximal token activation occurs on an irrelevant background region patch, its iERF correctly pinpoints the semantically meaningful evidence (spilled-pills) as the true attributional driver.
    This visualization demonstrates how CAFE resolves the ambiguity of non-localized features by grounding abstract activations in verifiable pixel-level evidence.}
    \label{fig:cafe_overview}
\end{figure*}
Let $A(p\mid z_k,I)$ denote the attribution of input patch $p$ to the scalar $z_k(I)$. 
We define the \emph{instance-specific Effective Receptive Field~(iERF)} of feature $k$ on image $I$ as
\begin{equation}
    \mathrm{iERF}_k(I) \;=\; \bigl\{\,\bigl(p,\,A(p\mid z_k,I)\bigr):\; p\in I\,\bigr\},
\end{equation}
whose intensity quantifies \yr{the attributional contribution of each patch, allowing us to identify the semantic driver even when it is spatially decoupled from the firing token.} 

As shown in Figure ~\ref{fig:cafe_overview}, we backpropagate relevance from the target SAE neuron through the SAE encoder and the vision transformer backbone. 
Within attention blocks, we employ \textbf{Attention-LRP (AttnLRP)}~\cite{achtibat2024attnlrp} to conserve and distribute relevance along attention edges. 

\subsection{Experiments}
\label{sec:cafe_experiments}
We conduct a series of experiments to evaluate the effectiveness of the proposed Concept-Anchored Feature Explanation (CAFE) framework.
Specifically, we evaluate whether iERF-based \emph{Concept-Anchored Feature Explanation}~(CAFE) (i) faithfully identifies the attributional evidence behind SAE activations, and (ii) clarifies when and where \emph{non-local} SAE features arise in a Vision Transformer~(ViT).

\paragraph{Experimental Setting}
All experiments use the CLIP--ViT-L/14 encoder of \cite{cherti2023reproducible}. 
For each transformer layer, we train a Matryoshka SAE \cite{huben2024sparse} on $5\times 10^{8}$ image patches from ImageNet-1K. 
Unless noted otherwise, reconstruction and sparsity hyperparameters are held fixed across layers.

\subsubsection{Quantitative Faithfulness Evaluation}
\label{subsubsec:insertion}
To assess the extent to which iERFs pinpoint the input evidence responsible for latent activations, we employ \emph{insertion} tests \cite{samek2016evaluating, achtibat2024attnlrp, han2024respect}. 
This procedure serves as a proxy for measuring attributional sufficiency by observing how rapidly the original activation $z_k$ is recovered as patches are iteratively inserted onto a blank \ssy{canvas} in descending order of importance.
We compare iERF-guided CAFE against a \emph{naive activation-ranking} baseline (top-$z_k$ patches) and several attributors used within \nj{CAFE} (KernelSHAP, AttnLRP, Integrated Gradients, and raw Gradients). 
We summarize performance by the area under the insertion curve (AUC) over images and features.

\begin{figure*}[ht]
    \centering
    \includegraphics[width=\textwidth]{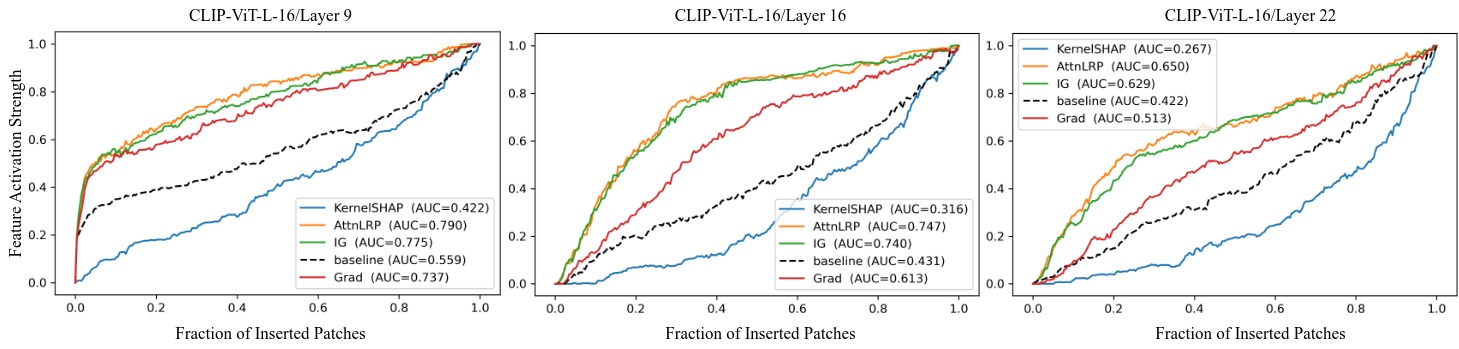}
    \caption{\textbf{Quantitative validation of attributional faithfulness.} 
    We compare iERF-guided CAFE with attribution variants (KernelSHAP, AttnLRP, Integrated Gradients, Gradients) against a naive activation-ranking baseline. 
    The CAFE instantiation with AttnLRP achieves the steepest insertion curves and the highest AUC, indicating more faithful identification of the input patches that satisfy the activation criteria of the latent.}
    \label{fig:insertion}
\end{figure*}

As shown in Fig.~\ref{fig:insertion}, iERF-guided CAFE generally outperforms activation-ranking across layers. 
Among the CAFE variants, AttnLRP yields the strongest gains (consistent with prior evidence of its fidelity for Transformers) \nj{followed closely by Integrated Gradients}. 

\subsubsection{Qualitative Observations of Non-localized Features}
\label{subsec:qualitative}
\begin{figure*}[ht]
    \centering
    \includegraphics[width=.95\textwidth]{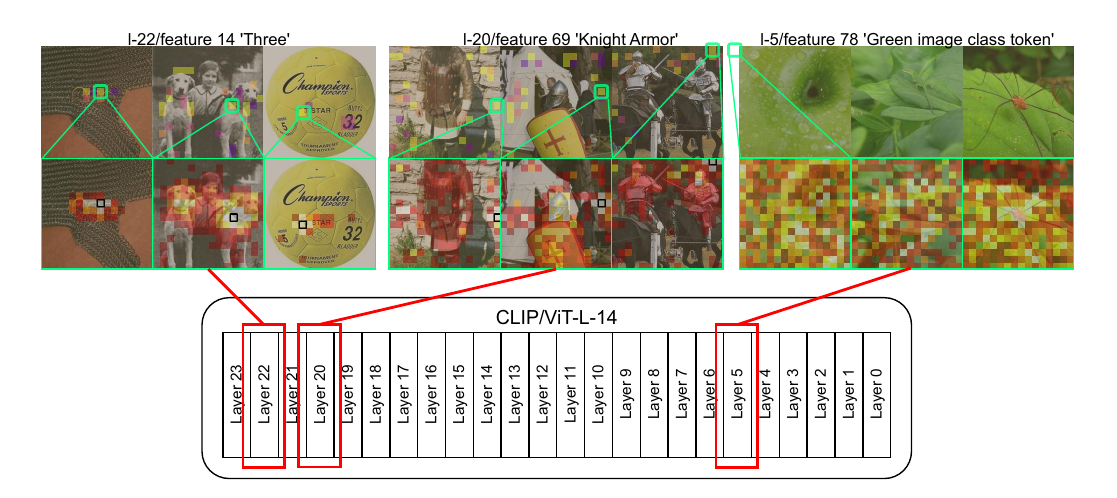}
    \caption{\textbf{Qualitative examples of non-localized SAE features and their iERFs across layers.}
    For each feature, we show the patches of maximal activation and the corresponding iERF. 
    Even when the maximal-activation tokens are spatially displaced from the region encoding the feature’s meaning, the iERF still pinpoints the true supporting evidence region. 
    In earlier layers, non-locality is rare and largely confined to class-token features; in later layers, globally mixed context makes non-local phenomena more common.}
    \label{fig:qualitative}
\end{figure*}
\yr{To complement the quantitative findings, we provide a qualitative analysis of non-localized SAE features, which traditional activation-ranking method \ssy{falters}.
As illustrated in \ssy{Fig.}~\ref{fig:qualitative}, we compare the baseline—which identifies features solely by their top-activated patches—with our CAFE framework.
The observations suggest that relying exclusively on activation magnitude can be misleading, as the firing tokens in deeper layers are often spatially decoupled from the semantic drivers that satisfy the feature's activation criteria.}

\yr{A notable example is found in \textit{Feature 14 of Layer 22}, which corresponds to the abstract concept of \textbf{``Three."} 
When inspecting the top-activated patches (baseline), the semantic consistency is difficult to discern because the tokens fire at seemingly arbitrary locations.
However, by anchoring these activations with iERFs, a clear ``counting" motif emerges: the iERF highlights three distinct circular objects in the first instance, a group of three entities (two dogs and one person) in the second, and the numerical digit ``3" on a ball in the third.
This indicates that the latent represents a high-level cardinality concept, a realization that remains hidden under activation-only inspection.} 

\yr{Similarly, in the case of \textit{Feature 69 of Layer 20}, \textbf{``Knight Armor"}, while the baseline patches show scenes containing armored knights, it remains ambiguous.
CAFE clarifies this ambiguity by demonstrating that the iERF is concentrated specifically on the metallic textures and structural patterns of the armor. 
This suggests that the latent is a specialized detector for armor primitives, resolving potential confounding where surrounding contextual cues might be mistaken for the primary activation driver.}

\yr{Furthermore, our framework provides a stable means to interpret \textbf{Class Token ([CLS]) features}, which are inherently challenging for previous patch-based methods.
Because the Class Token functions as a global summary and lacks a fixed spatial origin in the image plane, showing a ``top-activated patch" for a [CLS] feature is logically limited; it can only indicate which images the token prefers, but not why it does so spatially.
This ``spatial dimensionlessness" of the Class Token means that without a pixel-level trace, its semantic basis remains speculative.}

\yr{By employing iERFs, we can recover the pixel-level provenance of these abstract units.
For instance, in our \textbf{``Green Image"} Class Token feature, CAFE reveals that the token consistently aggregates evidence from green-hued regions across the entire scene. 
This capability to anchor spatially-decoupled or dimensionless tokens to verifiable input evidence underscores the necessity of iERF for a faithful interpretation of globally-mixed architectures like Transformers. 
As self-attention progressively mixes context, later-layer activations become increasingly difficult to interpret, a challenge that CAFE addresses by providing an explicit map of computational dependencies.}

\section{Mechanistic Interpretation: Interlayer Concept Graph}
\label{mechanistic}

\begin{figure*}[ht!]
    \centering
    \includegraphics[width=.9\linewidth]{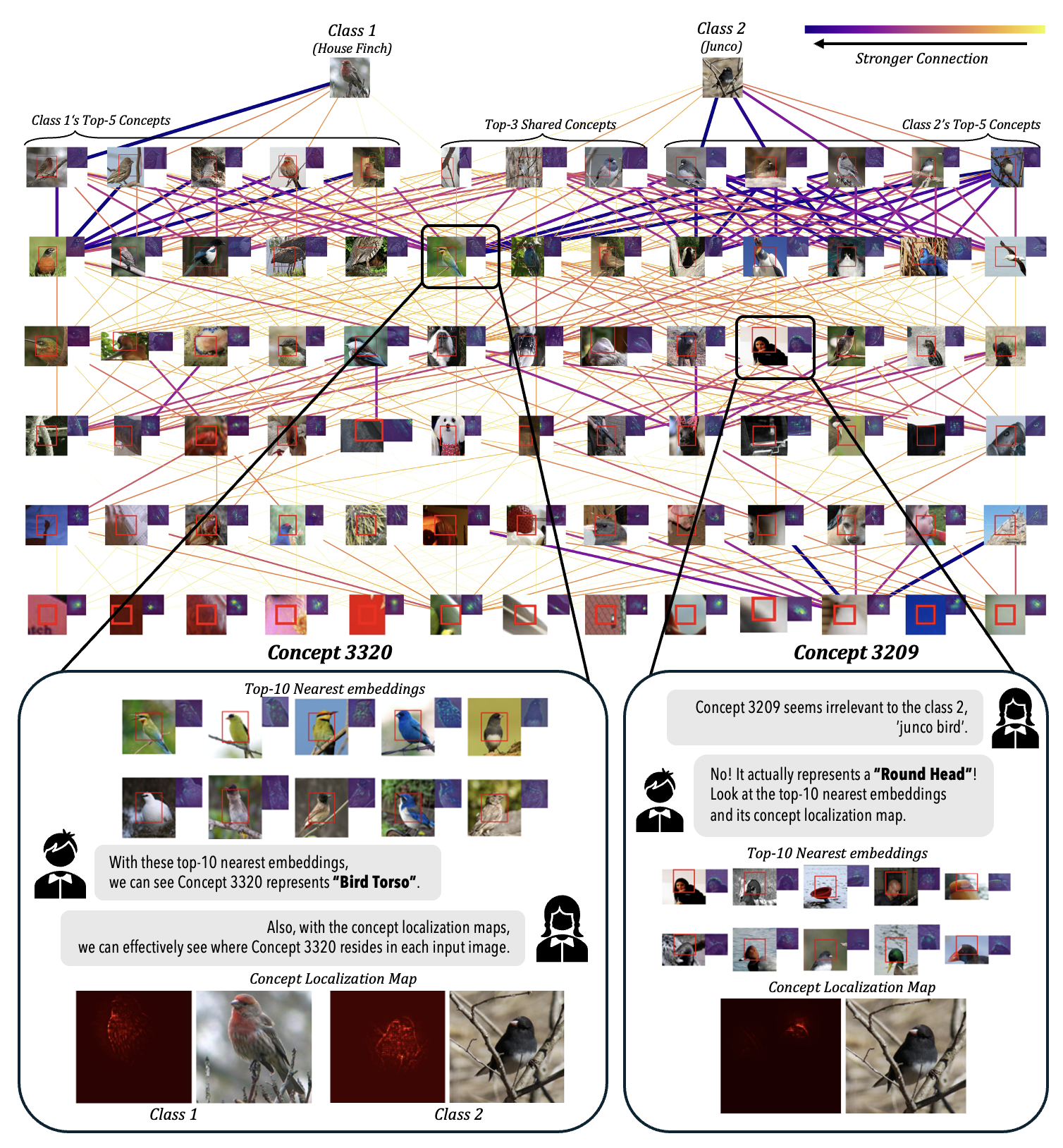}
    \caption{\textbf{Interlayer Concept Graph} of [Classifier, Layer4.2, Layer3.5, Layer3.2, Layer3.0, Layer2.3, Layer1.2], the bottleneck blocks in ResNet50.
    Nodes are iERF-anchored concept vectors (PFV–iERF bundles); directed edges quantify interlayer concept influence computed via ICAT.
    Edge width and color intensity (bluer = stronger) encode the contribution of a parent concept to its child. For clarity, we show the top-5 concepts per layer and the top-3 shared concepts; the full all-layer analysis appears in the supplementary materials.} 
    \label{fig: mechanistic_main}
\end{figure*}

\begin{figure*}[ht!]
    \centering
    \includegraphics[width=.85\linewidth]{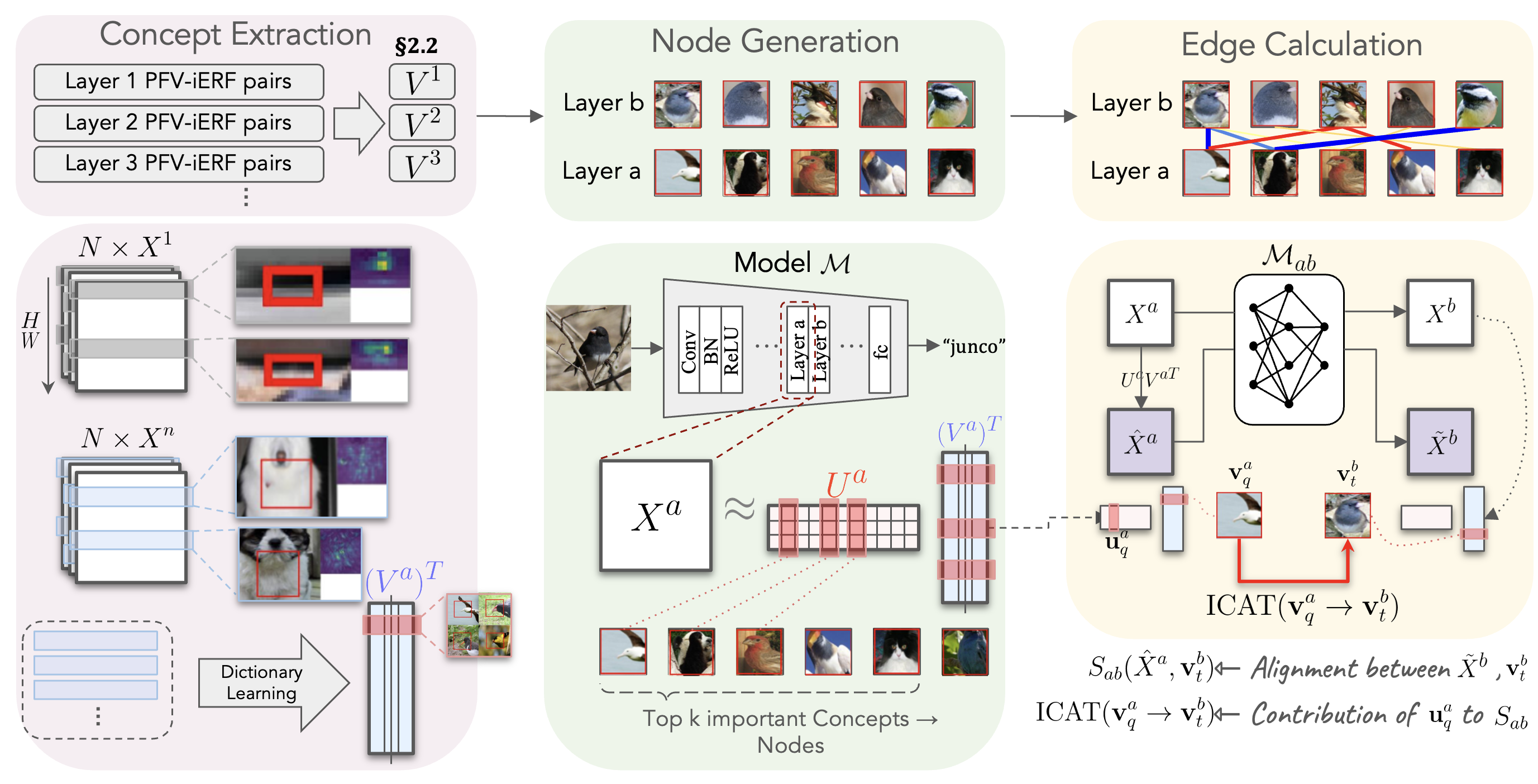}
    \caption{\textbf{Overview of our Interlayer Concept Graph~(ICG).} \textbf{Left.} Concept Extraction. The Pointwise Feature Vector (PFV) in the hidden layer is assigned a meaning by labeling it with the instance-specific Effective Receptive Field (iERF). Then, via dictionary learning, we extract concept matrix $V^l$ with gathered PFVs. 
    \textbf{Middle.} Node generation. We perform LASSO regression to find sparse concept coefficient matrix $U^l$, and select the top-$k$ important concepts as the nodes.
    \textbf{Right.} Edge Calculation. We calculate Interlayer Concept ATtribution (ICAT) of the query concept vector $\mathbf{v}_q^a$ to the target concept vector $\mathbf{v}_t^b$, where $a < b$.
    ICAT leverages an ATtribution method (AT) to determine importance of each concept coefficient ${\mathbf{u}_{q}^a}$ in computing concept alignment score $S_{ab}$.}
    \label{fig:method_overview}
\end{figure*}

\yr{While local and global explanations identify where evidence resides and what features are encoded}, mechanistic interpretability seeks to uncover \emph{how} internal computations assemble these elements into \nj{the behavior of the functional model}. 
In vision models, answering this requires tracing how intermediate \emph{concepts} are formed, combined, and transformed across layers into class evidence.
We tackle this by introducing \textbf{Interlayer Concept Graph~(ICG)}, a concept graph summarizing dominant computational pathways from input to output.
As shown in Fig.~\ref{fig: mechanistic_main}, the ICG enables the interpretation of how different model components relate to the final output and which specific parts of the input data are responsible for the evolution of these concepts across layers.
By anchoring each node with its iERF, we ensure that the entire mechanistic map remains grounded in pixel-level provenance rather than abstract activation magnitudes.
Fig.~\ref{fig:method_overview} overviews how it discovers layerwise concept vectors and traces concept pathways using \emph{Interlayer Concept Attribution}~(ICAT).
In brief, each layer is summarized by a compact concept dictionary in PFV space; edges quantify how concepts at layer $a$ contribute to concepts at layer $b>a$.
Because PFVs are labeled with iERFs, each concept inherits pixel-level provenance from its supporting PFVs.

\subsection{Nodes: Layerwise Conceptual Representations}
\label{sec: mechanistic_node}

The construction of the ICG begins by factorizing the Pointwise Feature Vector (PFV) space of each layer into a compact dictionary of concepts. 
We define a \emph{concept} as a (possibly overcomplete) linear basis in the PFV space that captures recurring semantics across images (textures, shapes, parts) beyond class boundaries~\cite{bricken2023monosemanticity}.

\subsubsection{Concept Extraction}
\label{extraction}
\begin{table*}[ht!]
    \centering
     \caption{Relative $\ell_2$ error and $\ell_{0}$ ratio for odd-numbered layers when reconstructing PFVs with different concept extractors. 
     For SAE, we used SAE encoder to get concept coefficients $U^l$. Dictionary learning was omitted for Layers 4.0 and 4.2 due to its enormous computational cost.}
    \resizebox{\textwidth}{!}{%
    \begin{tabular}{lr *{8}{r}} 
        \toprule
        \multirow{2}{*}{\textbf{Method}} & \multirow{2}{*}{} & \multicolumn{8}{c}{\textbf{Layers}} \\
        \cmidrule(lr){3-10}
                                         &  & \textbf{Layer1.1} & \textbf{Layer2.0} & \textbf{Layer2.2} & \textbf{Layer3.0} & \textbf{Layer3.2} & \textbf{Layer3.4} & \textbf{Layer4.0} & \textbf{Layer4.2} \\
        \midrule
        \multirow{2}{*}{Dictionary Learning} 
        & Rel-$l_{2}$(↓)  & 0.5968        & 0.6103        & 0.7795        & 0.6499        & 0.8235        & 0.7851        & -            & {-}     \\
        & $l_{0}$ ratio(↑)  & \textbf{0.9952} & 0.9958        & 0.9955        & 0.9953        & \textbf{0.9976} & \textbf{0.9982} & -            & -            \\
        \midrule
        \multirow{2}{*}{SAE} 
        & Rel-$l_{2}$(↓)   & 1.8952        & 1.5028        & 1.2524        & 1.2617        & 1.5257        & 1.4643        & \textbf{0.5096} & \textbf{0.4816} \\
        & $l_{0}$ ratio(↑) & 0.9622        & 0.9787        & 0.9779        & 0.9842        & 0.9889        & 0.9925        & 0.9921        & 0.9913        \\
        \midrule
        \multirow{2}{*}{Bisecting K-Means} 
        & Rel-$l_{2}$(↓)   & \textbf{0.3651} & \textbf{0.4633} & \textbf{0.6159} & \textbf{0.5618} & \textbf{0.6855} & \textbf{0.6565} & 0.6051        & 0.5095        \\
        & $l_{0}$ ratio(↑)  & 0.9947        & \textbf{0.9964} & \textbf{0.9961} & \textbf{0.9971} & 0.9967        & 0.9962        & \textbf{0.9968} & \textbf{0.9936} \\
        \bottomrule
    \end{tabular}%
    }
       \label{tab:PFVdecomposition}
\end{table*}

\sy{To obtain such bases, we evaluate three extractors—classical dictionary learning~\cite{efron2004least}, bisecting $k$-means clustering~\cite{karypis2000comparison}, and sparse autoencoders~\cite{bricken2023monosemanticity}—trained \emph{independently per layer} (i.e., concepts are not shared across layers).
For a layer $l$ with channel dimension $C_l$, we set the number of concepts to a fixed overcompleteness ratio, $K_l = 8C_l$, for all extractors to keep capacity proportional to layer width and to enable a fair comparison.
We compare these extractors using (i) relative reconstruction error, $\mathrm{Rel}\text{-}\ell_2 = \lVert\boldsymbol\epsilon\rVert_2/\lVert\mathbf{x}_p\rVert_2$, and (ii) sparsity, measured by the coefficient $\ell_0$ ratio, $1-\lVert\mathbf{u}_p\rVert_0/K_l$. For an illustrative overview of bisecting $k$-means and our use of cluster centroids as concept vectors, see Appendix~\ref{app:bisect_kmeans_pfv}.
} 

In the previous global analysis~(Section~\ref{sec:global_erf}), SAE were employed to prioritize the discovery of monosemantic latents. 
\sy{However, for a mechanistic trace that relies on faithful propagation across layers, high reconstruction fidelity is paramount, and extracted nodes should remain auditable via consistent supporting examples.
Recent work has also reported that standard SAE dictionaries can be unstable, and proposed constraining atoms to the data convex hull to improve stability and interpretability~\cite{fel2025archetypal}.
This motivates using extractors whose concepts are explicitly grounded in the empirical activation distribution; in particular, $k$-means concepts are computed as centroids of observed PFVs and thus are inherently data-anchored.}

As shown in Tab.~\ref{tab:PFVdecomposition}, bisecting $k$-means delivers the lowest reconstruction error across most layers while maintaining competitive sparsity; dictionary learning is marginally sparser in a few early blocks but at substantially higher error, and SAE tends to produce denser codes in intermediate layers.
We therefore adopt bisecting $k$-means as our default extractor \yr{for node generation in this mechanistic study}.
An illustrative qualitative comparison of exemplar coherence is provided in Appendix~\ref{appendix:qual_exemplar_coherence} (Fig.~\ref{fig:qual_kmeans_vs_sae}).
\subsubsection{Node Generation \yr{via Sparse Decomposition}}
\label{sec:node_generation}
At each layer, we decompose PFVs with the concept vectors.
Let there be $k$ concept vectors in a layer, denoted as $V := [\mathbf{v}_1, \cdots, \mathbf{v}_k] \in \mathbb{R}^{C \times k}$, residing in the $C$-dimensional vector space \( \mathcal{V} \) of the PFVs in the layer.
Each PFV $\mathbf{x}_p \in \mathbb{R}^C$ can be expressed as a linear combination of the concept vectors:
\begin{equation}
    \mathbf{x}_p = \sum_{j=1}^{k} u_{pj} \mathbf{v}_j + \mathbf{\epsilon} = V \mathbf{u}_p + \mathbf{\epsilon},
\end{equation}
where $u_{pj}$ is the coefficient representing the contribution of the $j$-th concept vector to PFV $x_p$ ($\mathbf{u}_p = [u_{p1},\cdots, u_{pk}]^T$), and $\epsilon$ is the residual error.
To determine the coefficients $\mathbf{u}_p$, we use Lasso regression~\cite{efron2004least}, which minimizes the following objective function:
\begin{equation}
    \mathbf{u}_p^* = \underset{\mathbf{u}_p}{\arg\min} 
    \left\{ \frac{1}{2} 
    \left\| \mathbf{x}_p 
    -V \mathbf{u}_p \right\|_2^2 
    + \lambda \left\| \mathbf{u}_p \right\|_1
    \right\},
\end{equation}
where $\lambda$ is a regularization parameter that controls the sparsity of the solution, encouraging many of the coefficients $u_{pj}$ to be zero.
Lasso regression was chosen to enforce sparsity, ensuring that each PFV is represented by only a few meaningful concept vectors.

In this way, the embeddings in the $l$-th layer, $X^l \in \mathbb{R}^{HW \times C}$, can be approximated by $k$ concept vectors as $\hat{X}^l = U^l (V^l)^T$, where $U^l \in \mathbb{R}^{HW \times k}$ and $V^l \in \mathbb{R}^{C \times k}$ are the coefficient and concept vector matrices, respectively. 
After decomposing $X^l$ into its approximation $U^l(V^l)^T$, 
we select the top-$k$ most important concepts as nodes based on their contribution to classification. 
The importance of a concept $\mathbf{v}^l_q$ is calculated as $\sum_{i}^{H^{l}W^{l}}AT(\mathbf{u}^l_{iq}, \text{logit}_c)$, where $\mathbf{u}^l_{iq}$ denotes the coefficients associated with concept $\mathbf{v}^l_q$, and $\text{logit}_c$ denotes the logit value for class $c$.
This sum represents the total spatial contribution of the coefficients corresponding to the concept $\mathbf{v}^l_q$.

\subsection{Edges: \sy{Quantifying Interlayer Concept Attribution (ICAT)}}
\label{edge}

In this section, we present \textbf{Interlayer Concept ATtribution~(ICAT)}, a method for quantifying the influence between concepts across layers.
Specifically, ICAT measures the \sy{influence} of the query concept vector $v^q_a$ at the source layer $a$ (\emph{parent node}) on the target concept vector $v^t_b$ at the target layer $b$ (\emph{child node}).
The following sections detail the operational procedure for computing these attributional edges.

\subsubsection{Interlayer Transformation Analysis}
Let \(\mathcal{M}\) denote our model under investigation, with \(a\) and \(b\) representing the source and the target layer, respectively.
We define \(X^l\) (\(l \in \{a,b | a < b\}\)) as the embeddings at layer \(l\), and our objective is to quantify the contribution of a query concept vector \(\mathbf{v}_q^a\) in layer \(a\) (\emph{parent node}) to the formation of a target concept vector \(\mathbf{v}_t^b\) in layer \(b\) (\emph{child node}).
We first isolate the transformation between layers $a$ and $b$ by copying the model \(\mathcal{M}\) between these layers, denoted \(\mathcal{M}_{ab}\).
This isolation ensures a focused examination of local interlayer mapping without \sy{attributing effects to layers outside this range.}

\subsubsection{Conceptual Approximation of Activations}
\yr{To trace the influence of specific concepts, we first represent the continuous hidden activations at source layer $a$ within our discovered conceptual basis.}
The activations at source layer \(a\), ${X}^a$, are then approximated by decomposing into a linear combination of concept vectors: 
\(\hat{X}^a=U^a \bigl(V^a\bigr)^T,\)
where \(V^a \in \mathbb{R}^{C^a \times k^a}\) is the concept vectors identified at layer \(a\), and \(U^a \in \mathbb{R}^{H^aW^a \times k^a}\) represents the corresponding coefficient maps.
This formation enables a structured representation of high-dimensional activations in terms of meaningful basis vectors.

\subsubsection{Forward Propagation and Semantic alignment}
The approximated activations $\hat{X}^a$ are then propagated through the sub-network \(\mathcal{M}_{ab}\) to evaluate the influence of the approximated activations on $\mathbf{v}^{b}_{t}$.
We denote the resulting embeddings as \(\tilde{X}^b :=\mathcal{M}_{ab}(\hat{X}^{a})\) in layer \(b\).
These resulting embeddings represent the transformed feature representations derived from the concept decomposition in layer \(a\).

To quantify their alignment with the target concept vector \(\mathbf{v}_t^b\), we introduce the \textbf{Concept Alignment Score} as:
\begin{equation}
      S_{ab}(\hat{X}^a, \mathbf{v}_t^b) \;=\; \sum_{i}^{H^{b}W^{b}} 
      \Bigl(u^b_{it} \cdot 
      \cos\bigl((\tilde{X}^b_{i,:})^T, \mathbf{v}_t^b\bigr)\Bigr).
    \label{eq:align}
\end{equation}
Here, \(u^b_{it}\) is the activation coefficient of \(\mathbf{v}_t^b\) at spatial location \(i\), and \(\cos(\cdot,\cdot)\) is the cosine similarity.
We adopt cosine similarity, as it focuses on directional alignment rather than magnitude, capturing the semantic relevance of the concept vector and avoiding distortions caused by varying activation scales.
Weighting by \(u^b_{it}\) further ensures that PFVs with higher activation of the target concept contribute more to the overall alignment score. 

\subsubsection{Aggregation of Interlayer Concept ATtribution~(ICAT)}
\yr{\sy{The final step is to quantify how each parent concept coefficient contributes to the target alignment score\ssy{.}}
We compute the ICAT \sy{edge weight} by aggregating the attributions of the query concept coefficients~\(u_{pq}^a\)--the coefficient of \(\mathbf{v}_q^a\) at position \(p\)--across all spatial locations \(p\):
\begin{equation}
    \mathrm{ICAT}(\mathbf{v}_q^a \rightarrow \mathbf{v}_t^b) = \sum_{p=1}^{H^a W^a} \text{AT}\bigl(u_{pq}^a, S_{ab}(\hat{X}^a, \mathbf{v}_t^b) \bigr)
\end{equation}
where $AT(\cdot)$ represents an attribution method.
\sy{This aggregated score summarizes the interlayer attribution from the query concept $\mathbf{v}^a_q$ to the target concept $\mathbf{v}^b_t$ under the chosen attribution operator $AT(\cdot)$.
}
By repeating this process for various concept pairs, we construct the \textbf{Interlayer Concept Graph}, providing a mechanistic map that reveals how visual evidence is progressively transformed and composed through the model’s internal hierarchy.
}

\subsection{Validation with Interlayer C-Insertion/Deletion}
\yr{To identify the most faithful attribution method for ICAT}, we introduce the \textbf{Interlayer Concept Insertion/Deletion}, which extends the concept attribution framework originally proposed by~\cite{fel2024holistic} to the interlayer setting.
This metric systematically inserts and deletes \textit{query concept vectors} in the source layer $a$ to observe their impact on the corresponding \textit{target concept vectors} in the subsequent layer $b$. 
\begin{figure*}[t]
    \centering
    \includegraphics[width=\linewidth]{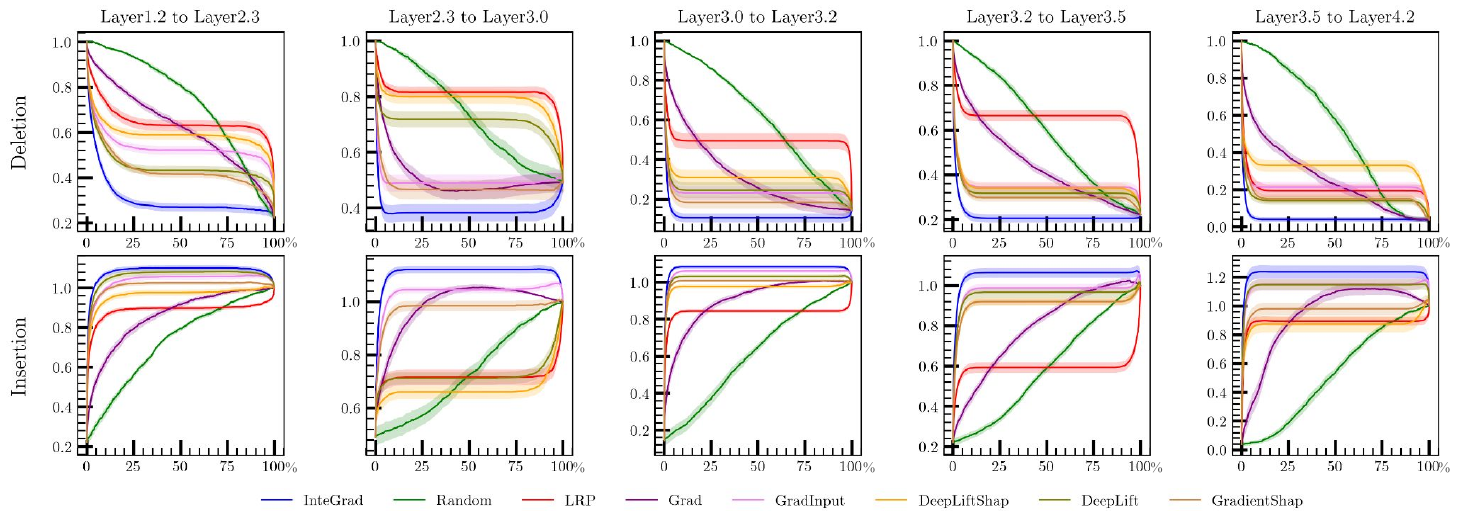}
    \caption{\textbf{Interlayer concept insertion/deletion on ResNet50.}
    Mean,$\pm$,SEM over 100 ImageNet-val images. We progressively delete or insert source-layer parent concepts in descending importance and track the maximal cosine alignment to the target concept (normalized to 1.0 at baseline). $\mathrm{ICAT}_{\text{IG}}$ (blue) yields the sharpest drop under deletion and the fastest rise under insertion, indicating the most faithful identification of influential parents.}
    \label{fig:Interlayer}
\end{figure*}
%

\subsubsection{Validation Protocol}
The validation procedure involves observing the response of a target concept~(child) $\mathbf{v}_t^b$ at layer $b$ when its supporting query concepts~(parent) in the preceding layer $a$ are modified.
In the \textbf{deletion} case, we delete the concept vectors from the approximated activations \(\hat{X}^{a}\) at layer \(a\) by setting their corresponding coefficients to 0.
The modified activations \(\hat{X'}^a\) are then propagated through the partial model \(\mathcal{M}_{ab}\), producing the resulting embeddings \(\tilde{X'}^b=\mathcal{M}_{ab}\bigl(\hat{X'}^a\bigr)\) at layer $b$.
We then measure the alignment between the target concept vector \(\mathbf{v}_t^b\) and each local embedding \(\tilde{x}_i^b\) (corresponding to the \(i\)-th spatial location in \(\tilde{X'}^b\)) by computing $\cos\bigl(\tilde{x}_i^b,\;\mathbf{v}_t^b\bigr)$.
We aggregate these alignments by taking the maximum value among the local alignments to mitigate noise. 

As we incrementally insert or delete concepts in the order of their importance, as indicated by an ICAT method, the alignment with the target concept vector in layer $b$ will increase (for insertion) or decrease (for deletion) more rapidly for the most critical concepts. 
If ICAT correctly identified important concepts, the rate of change will be steeper.
By plotting these changes, we can quantitatively assess how faithfully each ICAT method captures the \sy{most influential} concepts.

\subsubsection{Findings}
We applied to ResNet50, analyzing interlayer interactions across consecutive layers ([Layer1.2, Layer2.3, Layer3.0, Layer3.2, Layer3.5, Layer4.2]).
As demonstrated in Figure~\ref{fig:Interlayer}, Integrated Gradients (IG) consistently outperforms other attribution methods~\cite{shrikumar2016not, lundberg2017unified, shrikumar2019deeplift, bach2015LRP} in both interlayer insertion and deletion tasks, evidenced by its steeper decline in deletion and sharper rise in insertion.
This confirms the suitability of IG for tracing interlayer concept interactions, leading us to adopt it as the primary method for further analysis.
Interestingly, the insertion curves for IG sometimes exceed 1, implying that adding only positively attributed concepts can yield a stronger target activation than using all the 100\% of concept vectors. 
Once negatively attributed concepts are restored, the metric returns to 1.

\subsubsection{ICAT using Integrated Gradients}
Given its superior performance, we adopt Integrated Gradients (IG) as our main instantiation of the unified ICAT framework, referred to as $\mathrm{ICAT}_{IG}$.
Since $\mathrm{ICAT}_{IG}$ follows the idea of Integrated Gradients, the approximated activations \(\hat{X}^a\) are scaled from \(\alpha = 0\) to \(\alpha = 1\), allowing the integrated computation of gradients with respect to the defined alignment score:
\begin{equation}
  \text{AT}_{IG}\bigl(u^{a}_{pq}, S_{ab} \bigr)
  =
  u_{pq}^a 
  \int_{\alpha=0}^{1} 
  \frac{\partial S_{ab}(\alpha \hat{X}^a, \mathbf{v}_t^b)}
       {\partial\,u_{pq}^a}
  \, d\alpha.
\end{equation}
Here, $\text{AT}_{IG}$ is the attribution value using IG, \(S_{ab}(\cdot,\cdot)\) is the alignment score defined in Eq.~\ref{eq:align}, and \(u_{pq}^a\) is the activation strength of the query concept vector \(\mathbf{v}_q^a\) at location \(p\).

To compute the final attribution score, we sum over all spatial positions, \(p\):
\begin{equation}
\small
  \mathrm{ICAT}_{IG}(\mathbf{v}_q^a \rightarrow \mathbf{v}_t^b)
  =
  \sum_{p=1}^{H^a W^a}
  \text{AT}_{IG}\bigl(u^{a}_{pq}, S_{ab}(\hat{X}^a, \mathbf{v}_t^b) \bigr).
  \label{eq:icat_ig}
\end{equation}
This represents the overall contribution of the query concept vector $v^a_q$ to the target concept vector $v^b_t$, corresponding to the edge from the parent node $v^a_q$ to the child node $v^b_t$.

\subsection{Concept Tracing for Understanding Model}
\label{experiment}
In order to better understand how a model reaches its predictions, we construct a\ssy{n} \yr{interlayer concept} explanation graph (Fig.~\ref{fig: mechanistic_main}, with the full version in the supplementary materials). 
Each node in the graph is annotated with its top 10 nearest embeddings, indicating the regions where the concept is activated within an image.
These visual aids enable a thorough examination of how concepts are represented across diverse images (e.g., \emph{`bird torso'} vs. \emph{`round head'}) and where it appears in the original input.

Notably, our concept graph effectively captures both high-level features (e.g., object parts or semantic attributes) and low-level features (e.g., colors or textures) that might not be visually prominent.
For example, in Fig.~\ref{fig: mechanistic_main}, the concept of \emph{`red'} in a lower layer is crucial to identify the class \emph{house finch} despite its sparse presence relative to the dominant gray background.
Importantly, the final classification heavily depends on this seemingly rare feature, and our \yr{concept} graph illustrates how it propagates upward, contributing to more abstract representations such as \emph{`bird chest'} in higher layers. 

Furthermore, this approach provides a comprehensive visualization of the dependencies between concepts across different classes.
It explains not only which concepts are crucial for each class but also how shared concepts contribute to class similarities.
For example, visually similar classes may overlap on certain key features (e.g., \emph{`round head'} or \emph{`bird torso'} in Fig.~\ref{fig: mechanistic_main}), while maintaining unique attributes such as red plumage patterns distinguishing the class \emph{house finch}.
This level of interpretability surpasses conventional feature attribution methods or class-wise saliency methods, offering a more fine-grained, interpretable view of the model's internal representations.

\subsection{Diagnostic Utility of \yr{Interlayer Concept Graph}}

\begin{figure*}[t!]
    \centering    
    \includegraphics[width=.9\textwidth]
    {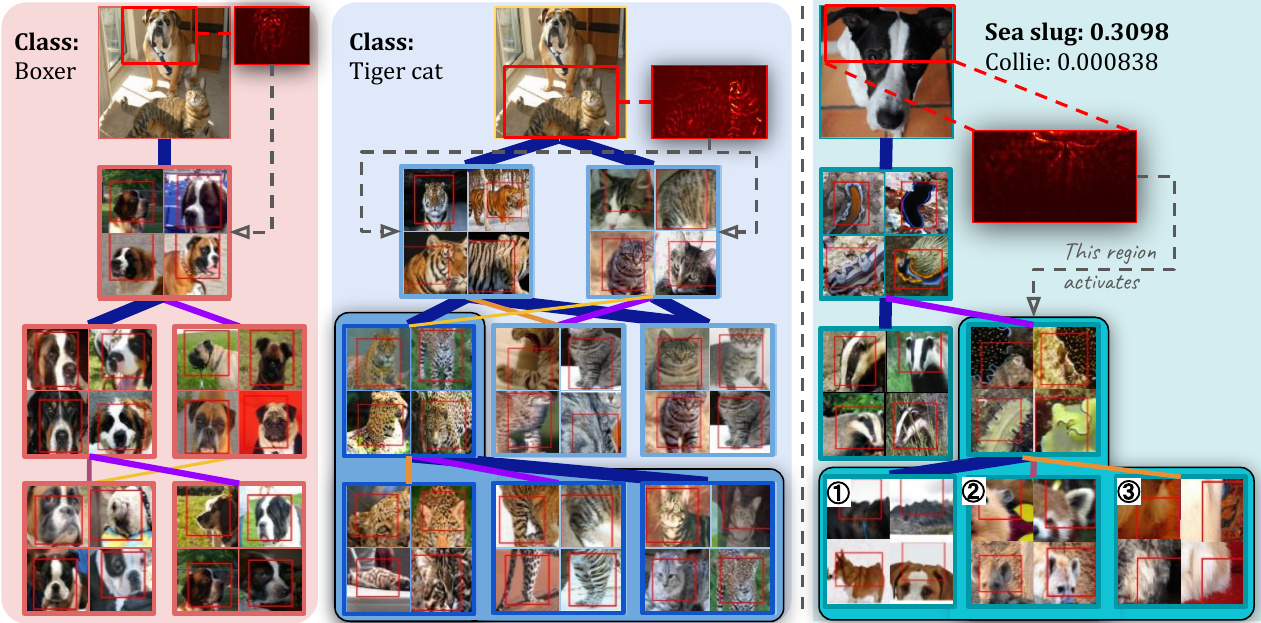}
    \caption{\textbf{Abbreviated Interlayer Concept Graphs} for two diagnostic cases (top-3 layers only).
    \textbf{Left}: Misclassification case. A \emph{tiger-cat} image is misclassified as \emph{boxer dog}, \yr{as a \emph{boxer dog} feature predominantly appears in the upper part of the image.
    \yr{Interlayer Concept Graph} diagnoses the confusion by showing the spurious red concept path leading to \emph{boxer}.}
    \textbf{Right}: Targeted attack case. After perturbation, a \emph{border collie} is re-labeled as \emph{sea slug} (final logits 0.00084 vs. 0.310); paths ①–③ show how low-level concepts such as \emph{`ear-shape'} and \emph{`wrinkle-shape'} combine into a corrupted higher-level \emph{`sea-slug'} concept.}    
    \label{fig:Experiments}
\end{figure*}

We qualitatively evaluated \yr{our} framework in two challenging scenarios, demonstrating its effectiveness in uncovering failure mechanisms in image classification models.

\subsubsection{Misclassification (\emph{tiger cat} $\rightarrow$ \emph{boxer dog})}
Our \yr{interlayer concept graph} reveals the pathways responsible for this misclassification by showing each \yr{concept} route where different parts of the image influence the final decision.
For example, in the left part of Fig.~\ref{fig:Experiments}, a ground-truth \emph{tiger cat} image is misclassified as \emph{boxer dog} due to the competing influence of a \emph{boxer dog} feature present in the upper part of the image.
There are two dominant \yr{concept} routes: the red route leading to class \emph{boxer dog} and blue of class \emph{tiger cat}.
The blue route, emerging from the lower half of the image, activates a chain of concepts leading to \emph{`tiger cat'} and \emph{`tiger'} concepts.
Interestingly, both concepts share an intermediate concept, with the only distinction being the edge strength of \emph{`striped feline'} concept.
\yr{Interlayer Concept Graph} enables us to trace how this erroneous intermediate concept formed by \emph{`striped feline face'}, \emph{`striped feline legs'}, \emph{`striped cat face'}, which naturally occur in \emph{tiger cat} images (see Fig.~\ref{fig:Experiments} blue box).
Concurrently, a red route from the upper half activates competing \emph{`boxer dog'} concepts, which ultimately lead to misclassification as class \emph{boxer}. 

\yr{Interlayer Concept Graph} effectively shows how the conflicting features—\emph{boxer dog} features and \emph{tiger cat} features—result in the misclassification. 
This highlights the model's inability to properly resolve the competing influences, allowing us to diagnose the source of the error.

\subsubsection{Targeted adversarial attack (\emph{border collie} $\rightarrow$ \emph{sea slug})}
The right part of Fig.~\ref{fig:Experiments} illustrates a Fast Gradient Sign Method (FGSM) attack \cite{goodfellow2014explaining} that forces a \emph{border collie} image to be misclassified as \emph{sea slug}.
\yr{Interlayer Concept Graph} reveals the hidden pathway exploited by the attack: the perturbation repurposes a strong \emph{`black-white stripe'} concept (originally tied to the dog’s fur) and injects a weak \emph{`sea-slug wrinkle'} concept.
Together, these concepts yield a higher-layer \emph{`sea-slug'} concept that overwhelms the original \emph{border-collie} representation.
Paths \textbf{①–③} trace how the corrupted high-level \emph{`sea-slug wrinkle'} concept emerges from lower-level concepts such as \emph{`wrinkle-like shape'}, \emph{`white-black rounded ear'}, and \emph{`white fur'}.
Although these individual concepts are not inherently corrupted, their combination results in the corruption of the final representation. 

In both scenarios, \yr{Interlayer Concept Graph} effectively pinpoints \emph{where} and \emph{how} erroneous or adversarially-induced concepts arise, providing a concrete handle for debugging misclassifications and assessing robustness.

\section{Conclusion}
We presented an iERF-centric framework that unifies \emph{local}, \emph{global}, and \emph{mechanistic} interpretability around a single analysis unit: the \textbf{pointwise feature vector (PFV)} labeled with its \textbf{instance-specific Effective Receptive Field (iERF)}. 
On the local side, \textbf{Sharing Ratio Decomposition (SRD)} constructs class-discriminative saliency by decomposing PFVs into their contributing receptive fields, yielding high-resolution and manipulation-resistant maps across architectures and activation functions. 
For the global view, we introduced \textbf{Concept-Anchored Feature Explanation (CAFE)} to explain \emph{non-localized} SAE features—particularly important in Transformers where early global mixing obscures pixel-level evidence. 
Finally, to answer \emph{how} representations are composed through depth, we proposed the \textbf{Interlayer Concept Graph} with \textbf{Interlayer Concept ATtribution (ICAT)}, which \ssy{quantifies} concept-to-concept influence across layers. 
Together, these components deliver a \yr{unified}, evidence-grounded map from pixels to concepts to decisions.

Empirically, SRD outperforms strong attribution and CAM baselines on fidelity, localization, sparsity, and robustness. 
CAFE recovers feature activations faster in insertion tests and reveals when non-localized SAE features emerge (rare in early layers, increasingly common in deeper Transformer blocks). 
The Interlayer Concept Graph exposes dominant concept routes behind correct predictions, misclassifications, and adversarial failures, enabling fine-grained debugging and hypothesis generation about model behavior.
\yyr{Positioned as an evidence-grounded interpretability pipeline, our framework provides a consistent attribution-based proxy for understanding how complex signals propagate through modern vision models.
Rather than claiming formal causal identification, it offers a systematic way to generate plausible explanatinos about internal computations—an approach supported by extensive empirical validation across diverse architectures.}
Grounded in iERFs, our approach links \emph{where} evidence arises, \emph{what} concepts are encoded, and \emph{how} they interact mechanistically—offering a coherent path toward transparent, trustworthy vision systems.

Looking ahead, we argue that interpretability should move beyond post-hoc visualization toward \emph{unified and concept-grounded} analyses that yield falsifiable predictions and support intervention. Future work includes extending our lens to generative, multimodal, and video models, strengthening \yr{attributional} identification with interventional probes, and turning explanations into levers for model steering—e.g., concept-level editing, ERF-regularized training, and dataset curation.


%

\bibliographystyle{IEEEtran}
\bibliography{main}

\begin{IEEEbiography}[{\includegraphics[width=1in,height=1.25in,clip,keepaspectratio]{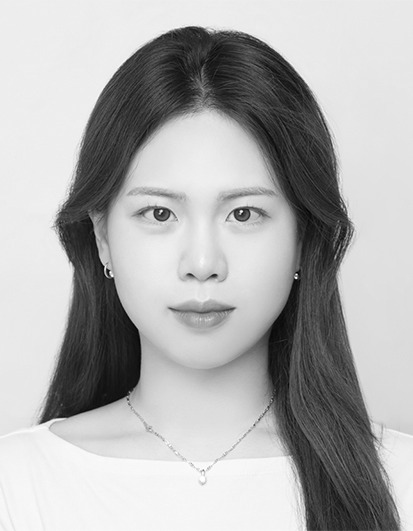}}]{Yearim Kim} received the B.S. degree (double major) in Statistics and Economics from Korea University, Seoul, Republic of Korea, in 2021. She is currently pursuing an integrated M.S./Ph.D. program in the Department of Intelligence and Information, Graduate School of Convergence Science and Technology, Seoul National University, Seoul, Republic of Korea.
Her research interests include Computer Vision and Explainable Artifical Intelligence~(XAI).
\end{IEEEbiography}

\begin{IEEEbiography}[{\includegraphics[width=1in,height=1.25in,clip,keepaspectratio]{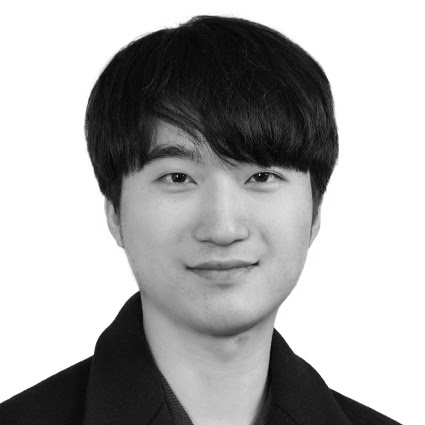}}]{Sangyu Han} received the B.S. degree in Health and Environment Science (with an interdisciplinary major in Artificial Intelligence) from Korea University, Seoul, Republic of Korea, in 2022. He is currently pursuing a combined M.S./Ph.D. in the Interdisciplinary Program of Artificial Intelligence at Seoul National University, Seoul, with research interests in explainable AI and the mechanistic interpretability of deep neural networks, including identifying monosemantic representations and developing robust attribution methods.
\end{IEEEbiography}

\begin{IEEEbiography}[{\includegraphics[width=1in,height=1.25in,clip,keepaspectratio]{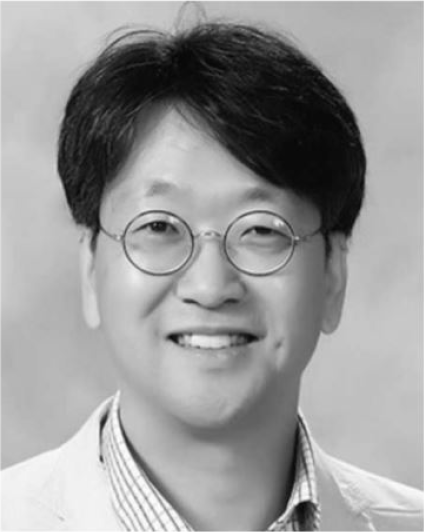}}]{Nojun Kwak} (Senior Member, IEEE) was born in Seoul, South Korea, in 1974. He received the B.S., M.S., and Ph.D.
degrees in electrical engineering and computer science from Seoul National University, Seoul, in
1997, 1999, and 2003, respectively. 
From 2003 to 2006, he was with Samsung Electronics, Seoul.
In 2006, he joined Seoul National University as a BK21 Assistant Professor. From 2007 to 2013, he was a Faculty Member with the Department of Electrical and Computer Engineering, Ajou University, Suwon, South Korea.
Since 2013, he has been with the Graduate School of Convergence Science and Technology, Seoul National University, where he is currently a Professor.
His current research interests include feature learning by deep neural networks and their applications in various areas of pattern recognition, computer vision, image processing, and natural language processing.
\end{IEEEbiography}

\vspace{11pt}

\clearpage
\section*{Appendix}
\subsection{Backward Pass: Estimating Sharing Ratios}
\label{local: backward}

\begin{figure*}[t]
\begin{center}
\includegraphics[width=.8\linewidth]{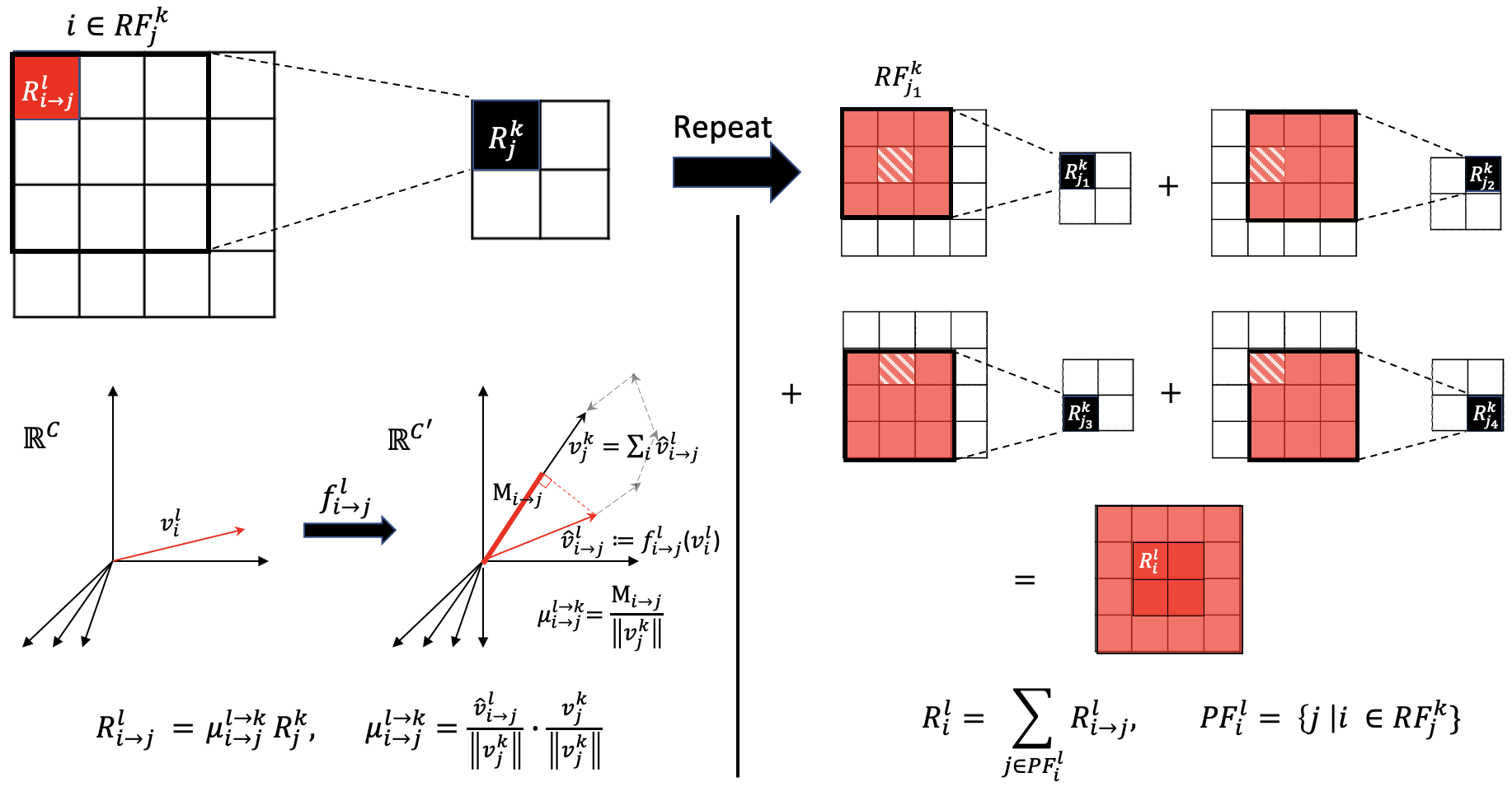}
\end{center}
\caption{Backward Pass of our method. $i$ and $j$ are pixels in activation layer $l$ and $k$, respectively. 
\textbf{Left}: 
$v^{k}_{j}$ is a pre-activation PFV at activation layer $k$, $v^l_i$ is a post-activation PFV at activation layer $l$, $f^{l}_{i \rightarrow j}$ is an affine transformation function assigned to $(i, j)$. 
Summation of every $\hat{v}^l_{i \rightarrow j}$ leads to $v_j^{k}$ ($\sum_{i \in RF_j^{k} } \hat{v}^l_{i \rightarrow j} = v_j^{k}$). $\mu^{l \rightarrow k}_{i\rightarrow j}$ is a sharing ratio of each $v^l_{i\rightarrow j}$ to $v_j^{k}$. 
$R_{i\rightarrow{j}}^{l}$ is the relevance share of $i$ in the leading layer to $j$ in the following layer. 
\textbf{Right}: $RF^{k}_j$ is the receptive field of pixel $j$ and $R^{k}_j$ is the relevance score of $j$ to the output. 
Relevance $R^l_i$ in the leading layer can be calculated recursively using the next layer's relevance $R^{k}_j$'s via $R^l_{i\rightarrow j}$'s for $j$'s whose receptive field includes pixel $i$.}
\label{fig:method_bp}
\end{figure*}

Figure~\ref{fig:method_bp} shows the backward pass of our sharing ratio decomposition.

Suppose a PFV $v_j^{k}$ is located at position $j$ immediately prior to the activation layer $k$. 
In a standard feed-forward network, the activation of $v_j^{k}$ is entirely determined by the PFVs in the preceding layer $l$ that fall within its receptive field, denoted as $RF_j^{k}$.
Formally,
\begin{equation}
v_j^{k} = f(V_j^{kl}) = \sum_{i \in RF_j^{k}} f_{i \rightarrow j}^{l}(v_i^l) 
= \sum_{i \in RF_j^{k}} \hat{v}_{i \rightarrow j}^l,
\end{equation}
where $V_j^{kl} = \{v_i^l \mid i \in RF_j^{k}\}$, and $f(\cdot)$ represents an affine transformation. 
Thus, each PFV $v_j^{k}$ can be decomposed into partial contributions $\hat{v}_{i \rightarrow j}^l$, each determined solely by a single PFV $v_i^l$ from the preceding layer.

We define the relevance $R_j^{k}$ of PFV $v_j^{k}$ as its contribution to the model output, typically a logit. 
To propagate relevance to the previous layer, we introduce the \emph{sharing ratio} $\mu^{l \rightarrow k}_{i \rightarrow j}$, which quantifies the relative contribution of PFV $v_i^l$ to $v_j^k$. 
This ratio is computed by taking the inner product between $\hat{v}_{i \rightarrow j}^l$ and $v_j^k$, normalized by the magnitude of $v_j^k$:
\begin{equation}
\mu_{i\rightarrow j}^{l\rightarrow k} = 
\Big\langle \frac{\hat{v}_{i \rightarrow j}^{l}}{\lVert v_j^{k} \rVert}, 
\frac{v_j^{k}}{\lVert v_j^{k} \rVert} \Big\rangle, 
\qquad \text{with} \quad
\hat{v}_{i \rightarrow j}^l = f_{i \rightarrow j}^{l}(v_i^l),
\end{equation}
ensuring that
\begin{equation}
\sum_{i \in RF_j^k} \mu_{i \rightarrow j}^{l \rightarrow k} = 1.
\label{eq:sharing_ratio}
\end{equation}

Using the sharing ratio, the relevance of $v_j^k$ can be decomposed and redistributed to the contributing PFVs of the preceding layer:
\begin{equation}
R_{i \rightarrow j}^{l} = \mu_{i \rightarrow j}^{l \rightarrow k} R_j^{k}, 
\qquad \text{so that} \qquad 
R_j^{k} = \sum_{i \in RF_j^k} R_{i \rightarrow j}^{l}.
\label{eq:r-score}
\end{equation}

Finally, the relevance of each PFV $v_i^l$ is obtained by accumulating its contributions across all projective fields $PF_i^l$ that include it:
\begin{equation}
R_i^l = \sum_{j \in PF_i^l} R_{i \rightarrow j}^{l}, 
\qquad PF_i^l = \{j \mid i \in RF_j^{k} \}.
\end{equation}

The recursion is initialized at the last encoder layer $L$, where the relevance of PFV $v_i^L$ to class $c$ is defined directly from the modified sharing ratio in Eq.~\ref{eq:mu_last}:
\begin{equation}
R^{L}_{i \rightarrow c} = \mu^{L \rightarrow O}_{i \rightarrow c},
\end{equation}
representing the class-specific contribution of PFV $v^L_i$ to the output logit $y^c$.  
This initialization ensures that the backward propagation of relevance remains consistent with the class-discriminative refinement introduced in the forward process, thereby establishing a direct link between the two perspectives.

\subsection{Proof of Equivalence Between Forward and Backward Formulations}
\label{sec:equivalence}

We present here the detailed derivation establishing the equivalence of the forward and backward formulations introduced in Sec.~\ref{local}. 

\paragraph{Forward process}
The saliency map for class $c$ is given by
\begin{equation}
    \phi_c(x) = \sum_{i}\mu_{i \rightarrow c}^{L \rightarrow O} \cdot ERF_{v_i^L},
\end{equation}
where each $ERF_{v_i^L}$ can be recursively expanded as
\begin{equation}
    ERF_{v_i^{l+1}} = \sum_{j}\mu_{j \rightarrow i}^{l \rightarrow l+1} \cdot ERF_{v_j^l}.
\end{equation}
By repeatedly unrolling across layers, we obtain the path-based decomposition
\begin{equation}
    ERF_{v_i^L} = \sum_{p \in [HW]} \left( \sum_{\tau \in T} \prod_{l \in [L]} \mu^{(l-1)\rightarrow l}_{p_{l-1}\rightarrow p_l} \cdot E^p \right),
\end{equation}
where $\tau = (p_0=p, p_1, \cdots, p_{L-1}, p_L=i)$ is a trajectory from input pixel $p$ to position $i$ at layer $L$.  
Consequently, the saliency map becomes
\begin{equation}
    \phi_c(x) = \sum_{p}\sum_{i}\mu_{i \rightarrow c}^{L \rightarrow O}\left( \sum_{\tau \in T} \prod_{l\in [L]} \mu^{(l-1)\rightarrow l}_{p_{l-1}\rightarrow p_l} \cdot E^p \right).
    \label{eq:saliency_forward}
\end{equation}

\paragraph{Backward process}
The saliency map can equivalently be defined as the weighted combination of input basis vectors:
\begin{equation}
    \phi_c(x) = \sum_{p} R_p^0 \cdot E^p,
\end{equation}
with relevance scores propagated recursively:
\begin{equation}
    R_i^{l-1} = \sum_{j \in PF_i} \mu_{i \rightarrow j}^{(l-1)\rightarrow l} \cdot R_j^l.
\end{equation}
Unrolling this recursion yields
\begin{equation}
    R_p^0 = \sum_{i} R_i^L \left( \sum_{\tau \in T} \prod_{l \in [L]} \mu^{(l-1)\rightarrow l}_{p_{l-1}\rightarrow p_l} \right).
\end{equation}
Since $R_i^L = \mu_{i \rightarrow c}^{L \rightarrow O}$ by definition, the resulting saliency map is
\begin{equation}
    \phi_c(x) = \sum_{p}\sum_{i}\mu_{i \rightarrow c}^{L \rightarrow O} \left( \sum_{\tau \in T} \prod_{l\in [L]} \mu^{(l-1)\rightarrow l}_{p_{l-1}\rightarrow p_l} \cdot E^p \right),
    \label{eq:saliency_backward}
\end{equation}
which is identical to Eq.~\ref{eq:saliency_forward} derived from the forward process.

\paragraph{Equivalence}
The equivalence between Eqs.~\ref{eq:saliency_forward} and \ref{eq:saliency_backward} formally demonstrates that forward and backward formulations in our SRD framework are mathematically consistent, thereby validating the robustness of our interpretability method. 

\subsection{Ablation study on final sharing-ratio refinement}
\label{appendix:mu_ablation}
This section analyzes the refinement used to compute the final sharing ratios
$\mu^{L \rightarrow O}_{i \rightarrow c}$ from PFV-level class contribution scores $\Phi_i^c$
(Eq.~\ref{eq:mu_last}).
We compare the following variants:
(i) \textbf{Raw}, which directly uses $\Phi_i^c$ as the aggregation weight (no class-centering, no clamping);
(ii) \textbf{Mean}, which applies class-centering by subtracting the mean score across classes at the same PFV location
but keeps signed values (class-centering only; signed);
and (iii) \textbf{Clamped mean} (ours), which additionally takes the positive part after centering
(class-centering + positive-part).

\begin{figure}[]
\begin{center}
\includegraphics[width=\linewidth]{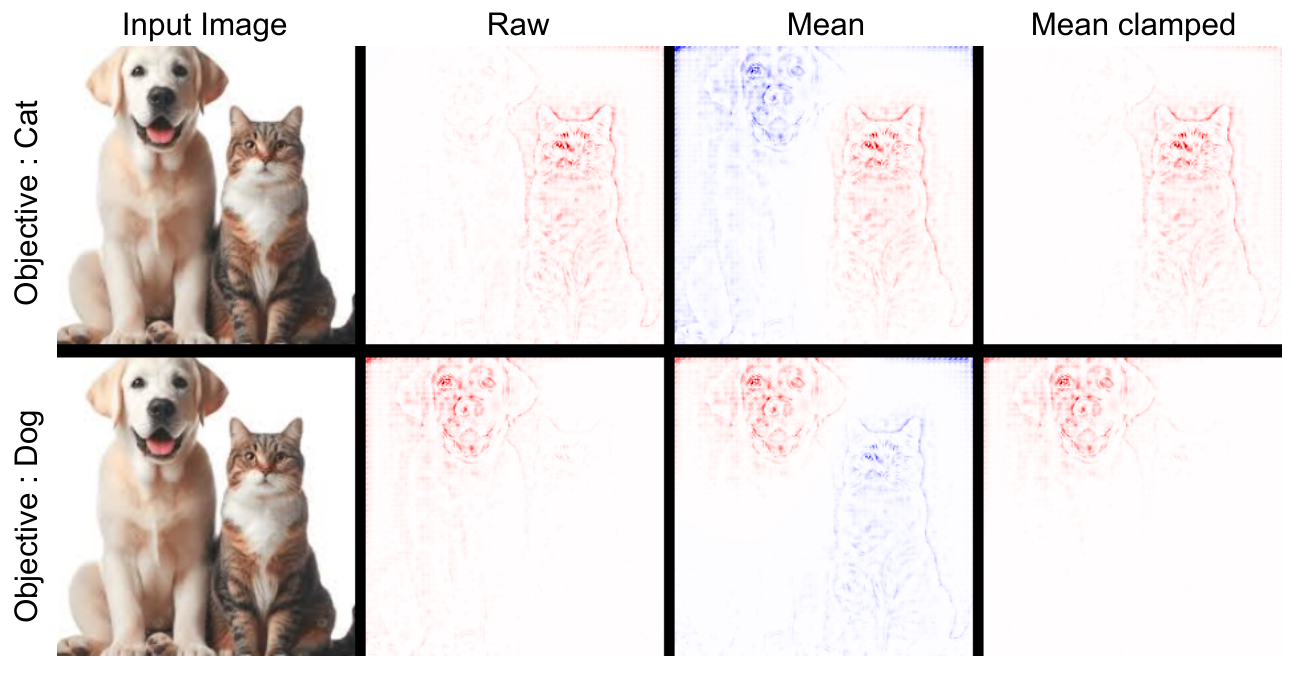}
\end{center}
\vspace{-5mm}
\caption{\textbf{Qualitative comparison of sharing-ratio variants.}
An example input contains both a cat and a dog.
Rows indicate the target class (\textit{Objective: Cat} / \textit{Objective: Dog}).
Columns show the input and saliency maps obtained with \textbf{Raw} ($\mu=\Phi$),
\textbf{Mean deducted} (class-centering only; signed),
and \textbf{Mean clamped} (ours; class-centering + positive-part).
The \textbf{Mean} variant produces signed maps where regions corresponding to competing classes can appear as
counter-evidence (e.g., negative responses), whereas \textbf{Mean clamped} suppresses such counter-evidence and yields a
cleaner positive-evidence map concentrated on the target object. }
\label{fig:mu_ablation_vis}
\end{figure}

Table~\ref{tab:mu_ablation} reports quantitative results on ImageNet-S50 (n=752) using two backbones.
Overall, \textbf{clamping after class-centering} substantially improves \emph{Attribution localization} and \emph{Sparseness} metrics
for both VGG16 and ResNet50, indicating better class-specific localization.
While class-centering alone (\textbf{Mean}) can increase pointing game (Poi.) in some cases,
it may also introduce prominent counter-evidence regions that degrade local attribution alignment.
The proposed \textbf{Clamped mean} variant yields more class-discriminative \emph{positive-evidence} maps,
which is consistent with the intended use of $\mu^{L \rightarrow O}_{i \rightarrow c}$ as aggregation weights
for saliency construction.
To illustrate the effect of clamping, we provide a qualitative comparison in Fig.~\ref{fig:mu_ablation_vis}.

\begin{table*}[]
\centering
\resizebox{\textwidth}{!}{%
\begin{tabular}{llllllllllllll}
\hline
             &  &  & \multicolumn{5}{c}{VGG16}                                                                                                                           & \multicolumn{1}{c}{} & \multicolumn{5}{c}{Resnet50}                                                                                                                        \\ \cline{4-8} \cline{10-14} 
             &  &  & \multicolumn{1}{c}{Poi.(↑)} & \multicolumn{1}{c}{Att.(↑)} & \multicolumn{1}{c}{Spa.(↑)} & \multicolumn{1}{c}{Fid.(↑)} & \multicolumn{1}{c}{Sta.(↓)} & \multicolumn{1}{c}{} & \multicolumn{1}{c}{Poi.(↑)} & \multicolumn{1}{c}{Att.(↑)} & \multicolumn{1}{c}{Spa.(↑)} & \multicolumn{1}{c}{Fid.(↑)} & \multicolumn{1}{c}{Sta.(↓)} \\ \hline
Raw          &  &  & .906                        & .490                        & .746                        & .068                        & .107                        &                      & .918                        & .507                        & .596                        & .075                        & .115                        \\
Mean         &  &  & .920                        & .404                        & .748                        & .071                        & .104                        &                      & \textbf{.958}                        & .439                        & .629                        & .078                        & .100                        \\
Clamped mean &  &  & \textbf{.925}                        & \textbf{.561 }                       & \textbf{.788}                        & .069                        & \textbf{.099}                      &                      & .953                        & \textbf{.576 }                       & \textbf{.724}                        & \textbf{.082     }                   & .104                        \\ \hline
\end{tabular}%
}
\label{tab:mu_ablation}
\caption{\textbf{Ablation of the final sharing-ratio refinement (Eq.~\ref{eq:mu_last}).}
We compare three variants for computing the encoder-output aggregation weights $\mu^{L\rightarrow O}_{i\rightarrow c}$ on ImageNet-S50 (n=752).
\emph{Raw}: no class-centering, no clamping.
\emph{Mean}: class-centering only; signed.
\emph{Clamped mean} (ours): class-centering + positive-part.
Higher is better for Poi./Att./Spa./Fid., and lower is better for Sta.}
\end{table*}

\subsection{Illustrative exemplar coherence comparison: SAE vs.\ bisecting $k$-means}
\label{appendix:qual_exemplar_coherence}
\begin{figure*}[t]
    \centering
    \includegraphics[width=\linewidth]{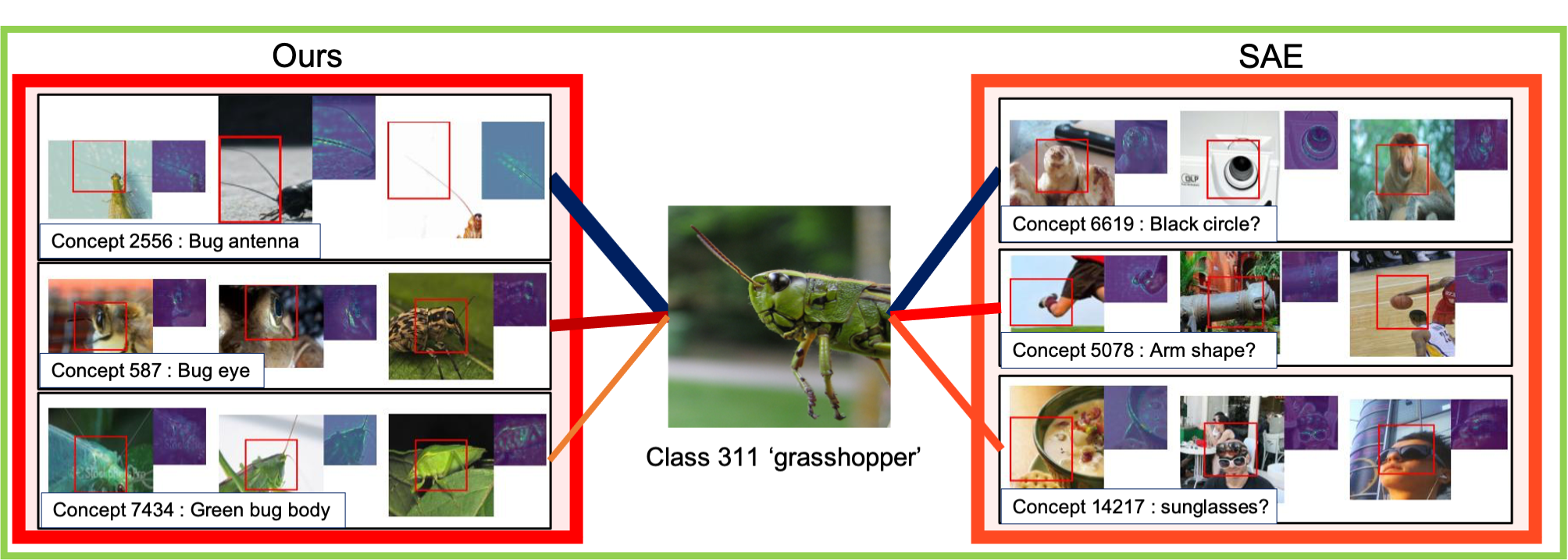}
    \caption{\textbf{Illustrative exemplar coherence for ImageNet class 311 (\textit{grasshopper}).}
    For each extractor (left: ours, bisecting $k$-means; right: SAE), we show three top-ranked concepts and their nearest-neighbor PFV/iERF exemplars.
    Red boxes indicate the evidence regions used for retrieval/visualization.
    In this example, the $k$-means concepts yield coherent, class-relevant evidence (e.g., antenna/eye/body cues), whereas the SAE concepts shown are harder to label from their exemplars.
    This figure is provided as a qualitative sanity check and is not used as the primary criterion for selecting the extractor.}
    \label{fig:qual_kmeans_vs_sae}
\end{figure*}

\label{app:qual_exemplar_coherence}
This appendix provides an illustrative qualitative sanity check on the auditability of extracted concepts (Fig.~\ref{fig:qual_kmeans_vs_sae}).
For a representative class (ImageNet class 311, \textit{grasshopper}), we compare the top-ranked concepts produced by SAE and bisecting $k$-means under the same concept-importance criterion used in our mechanistic pipeline.
For each concept, we retrieve nearest-neighbor PFV/iERF exemplars and visualize the corresponding evidence patches (Fig.~\ref{fig:qual_kmeans_vs_sae}).
In this example, bisecting $k$-means yields semantically coherent exemplars that are straightforward to label (e.g., antenna/eye/body-related evidence), whereas the SAE concepts shown are less consistently attributable to class-specific evidence.
We emphasize that this qualitative comparison is not used as a primary selection criterion; the extractor choice in the main text is based on the quantitative reconstruction/sparsity trade-off (Tab.~\ref{tab:PFVdecomposition}).

\subsection{Illustrative overview of bisecting $k$-means concepts}
\label{app:bisect_kmeans_pfv}

\begin{figure}[]
\includegraphics[width=0.9\linewidth]{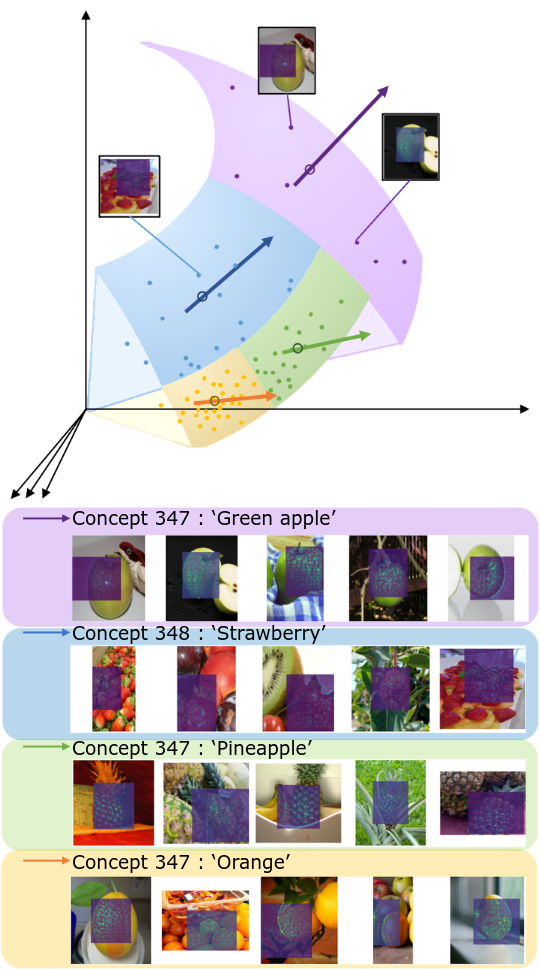}
\vspace{5mm}
\caption{\textbf{Illustrative overview of bisecting $k$-means on a non-uniform PFV distribution.}
\textbf{Top:} A schematic PFV space with density variations (dense vs. sparse regions). Bisecting $k$-means recursively partitions the PFV cloud into locally coherent clusters (shaded regions), and we use each cluster centroid as a concept vector (arrows).
\textbf{Bottom:} Nearest-neighbor exemplars retrieved around each centroid exhibit consistent evidence patterns, enabling concept auditability through exemplar-based inspection.}
\label{fig:bisect_overview}
\end{figure}

This appendix provides an illustrative overview of how we extract concepts using bisecting $k$-means in the PFV space (Fig.~\ref{fig:bisect_overview}).
The figure is intended as a schematic motivation and visualization of the procedure, rather than an empirical characterization of PFV density or geometry.

Bisecting $k$-means recursively partitions the set of PFVs within a layer into clusters by repeatedly splitting a cluster into two sub-clusters until the desired number of clusters is reached.
We represent each cluster by its centroid and use these centroids as concept vectors.
Because centroids are computed directly from observed PFVs, the resulting concept vectors are explicitly data-derived and can be audited via nearest-neighbor exemplar retrieval around each centroid.
Figure~\ref{fig:bisect_overview} visualizes this workflow and shows example retrieved evidence patches for several concepts.

\subsection{Additional analysis of local explanation}
\label{appendix: local_qualitative}
\subsubsection{Saliency map comparison}
\begin{figure*}[t]
\centering
\begin{minipage}[t]{.495\textwidth}
  \centering
  \includegraphics[width=\linewidth]{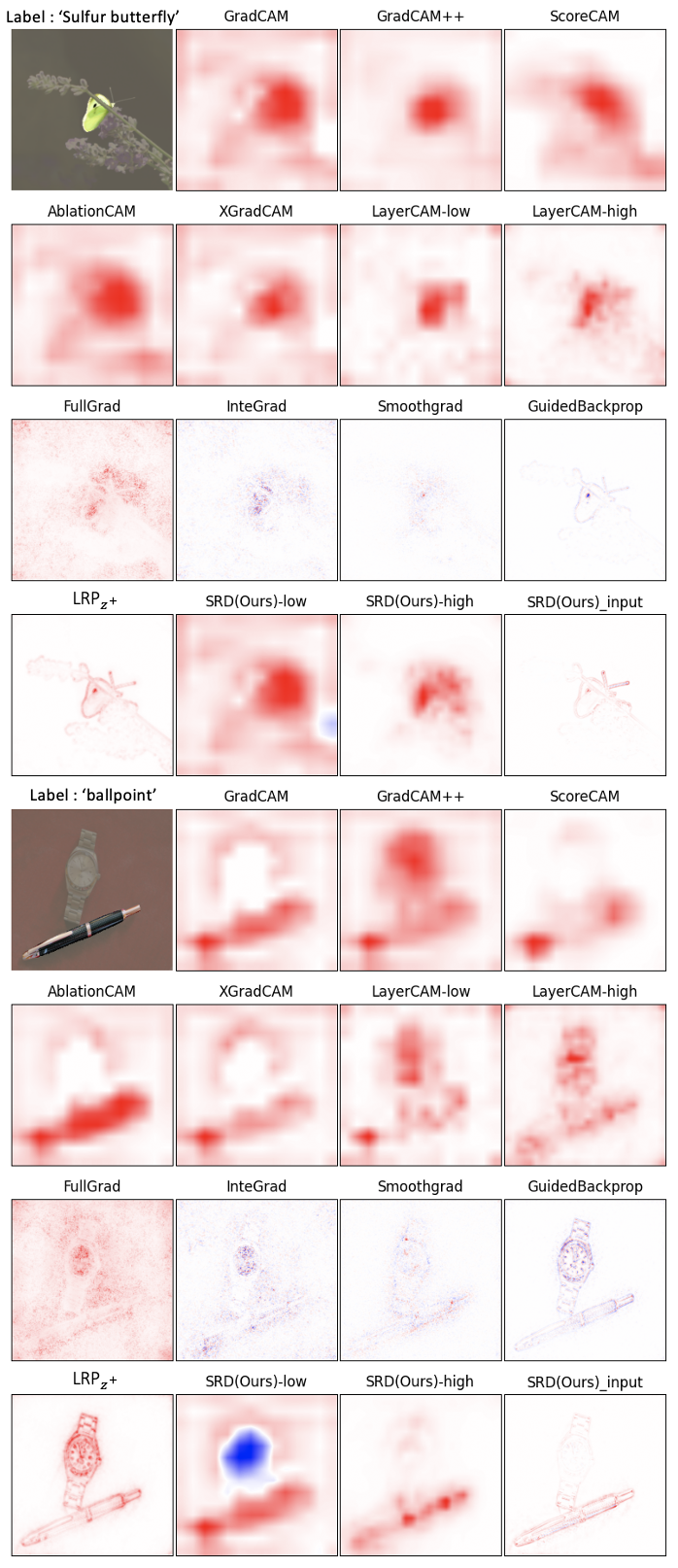}
\end{minipage}\hfill
\begin{minipage}[t]{.495\textwidth}
  \centering
  \includegraphics[width=\linewidth]{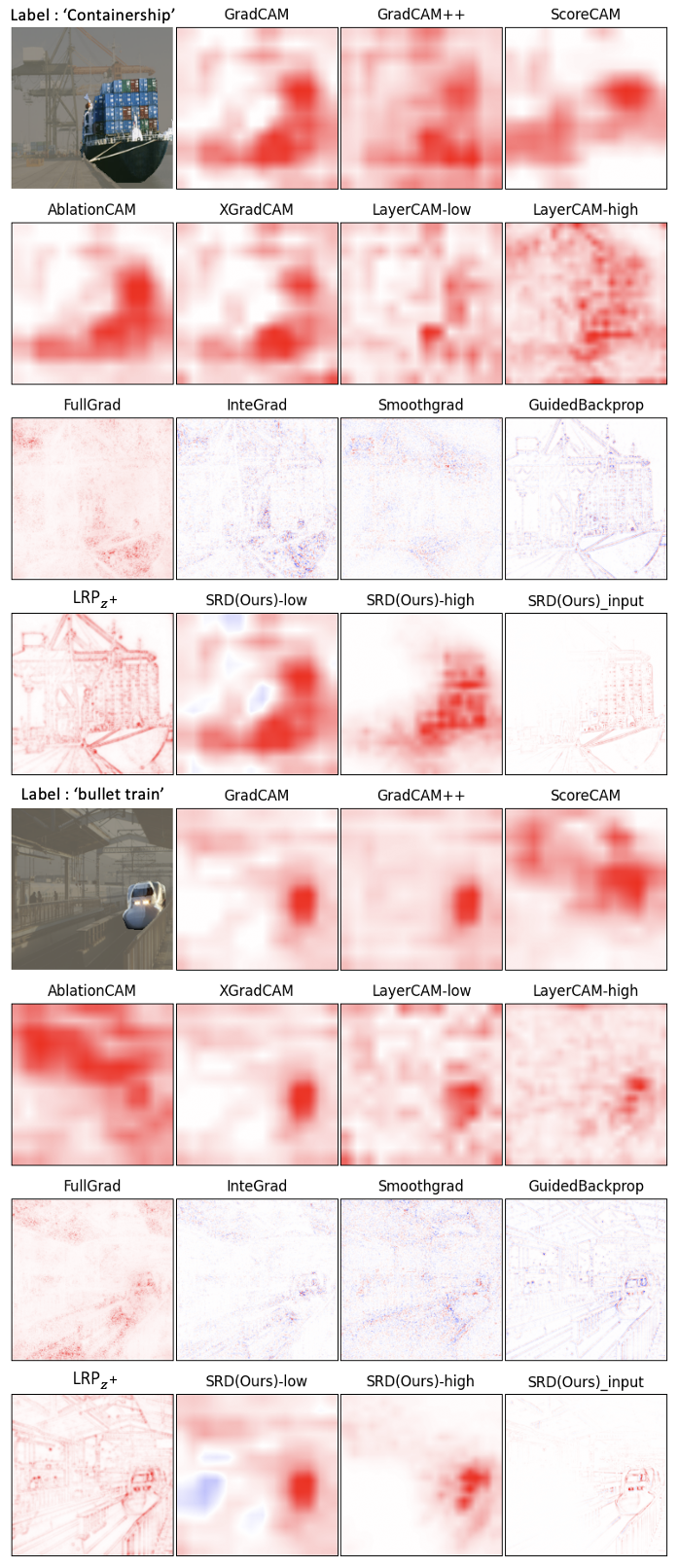}
\end{minipage}
\caption{Qualitative comparisons on VGG16. The highlighted region denotes the segmentation mask.}
\label{fig:saliency_vgg1}
\end{figure*}

\begin{figure*}[t]
\centering
\begin{minipage}[t]{.495\textwidth}
  \centering
  \includegraphics[width=\linewidth]{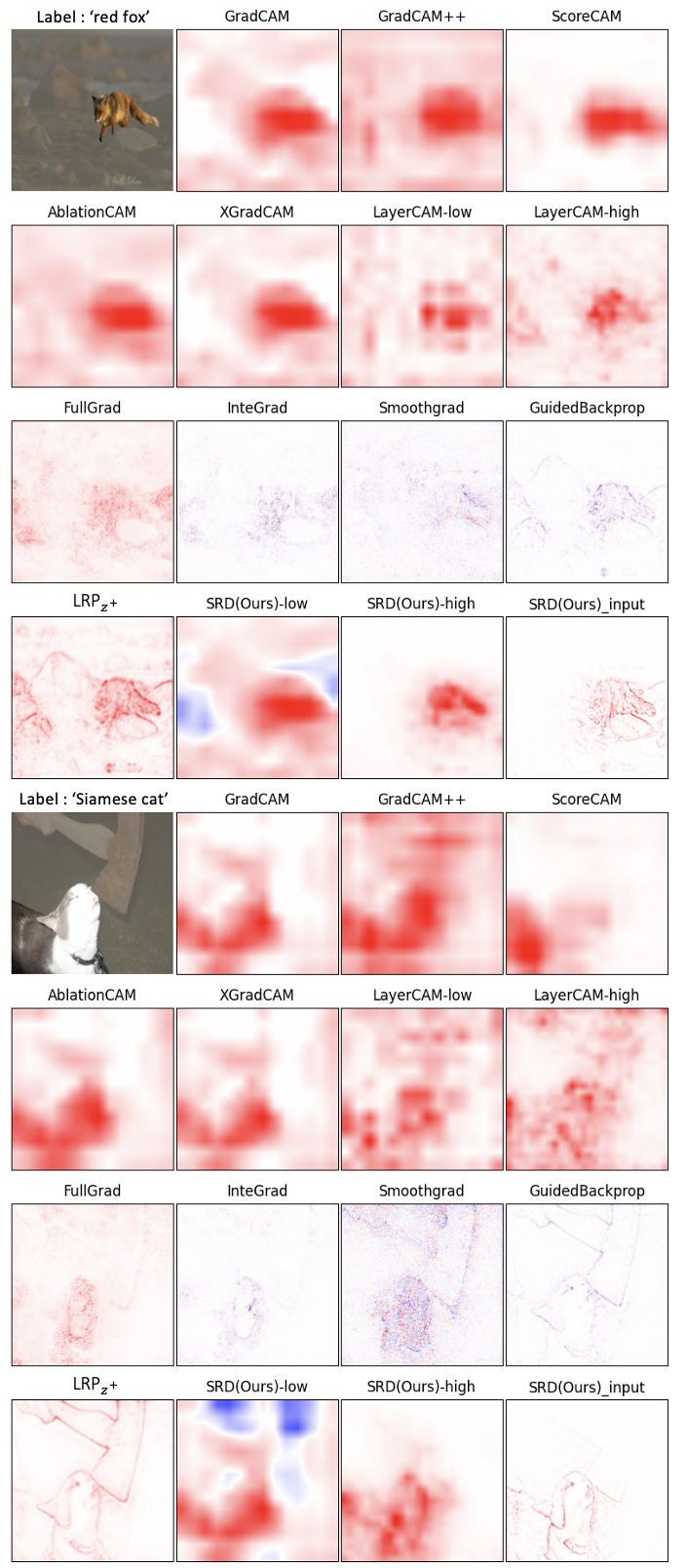}
\end{minipage}\hfill
\begin{minipage}[t]{.495\textwidth}
  \centering
  \includegraphics[width=\linewidth]{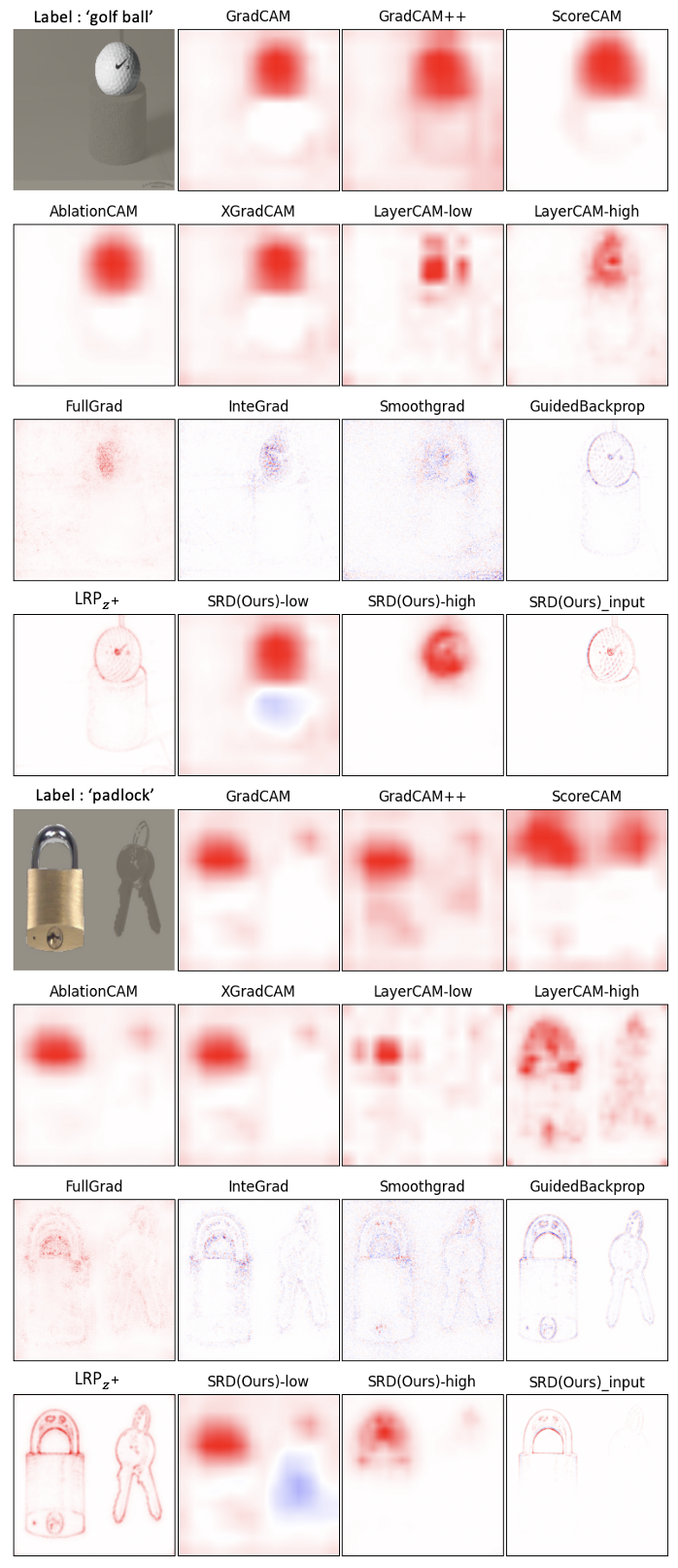}
\end{minipage}
\caption{Qualitative comparisons on VGG16. The highlighted region denotes the segmentation mask.}
\label{fig:saliency_vgg2}
\end{figure*}

\begin{figure*}[t]
\centering
\begin{minipage}[t]{.495\textwidth}
  \centering
  \includegraphics[width=\linewidth]{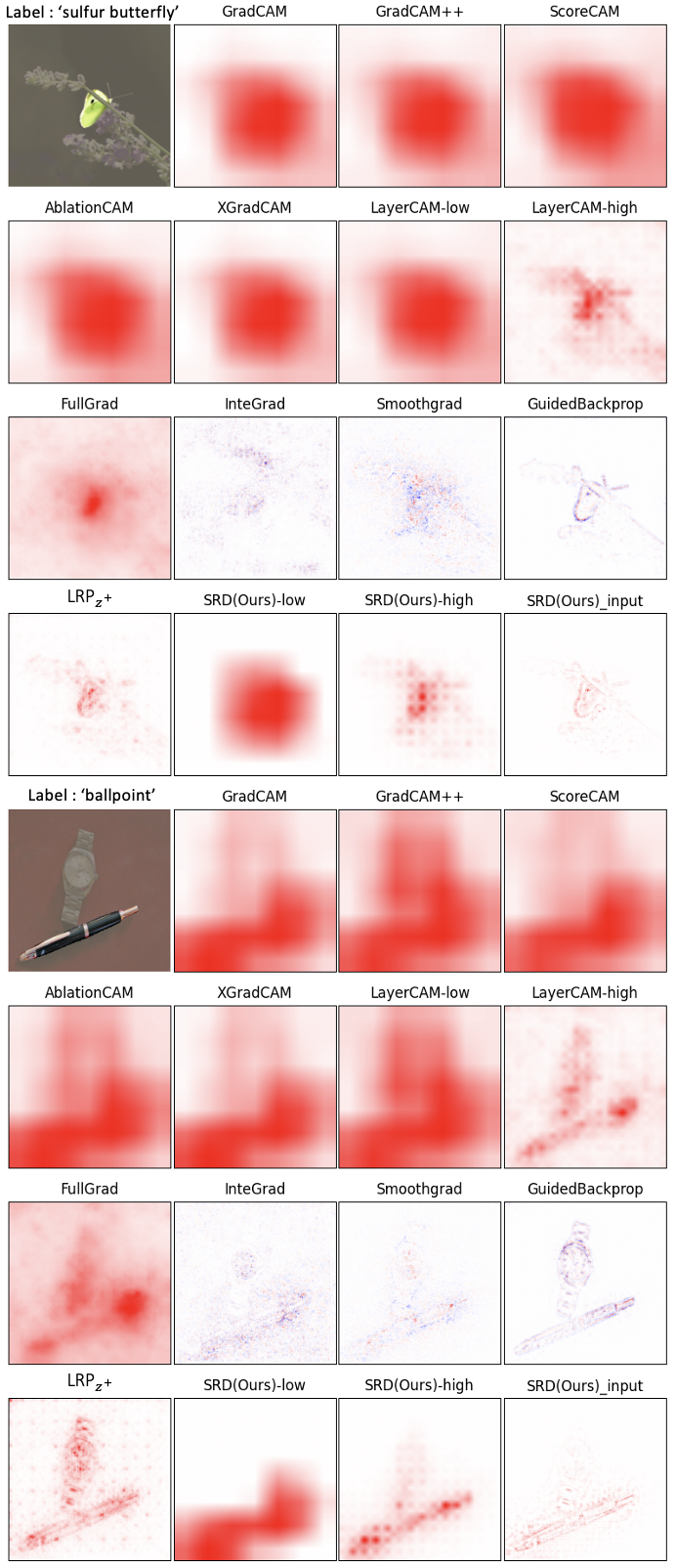}
\end{minipage}\hfill
\begin{minipage}[t]{.495\textwidth}
  \centering
  \includegraphics[width=\linewidth]{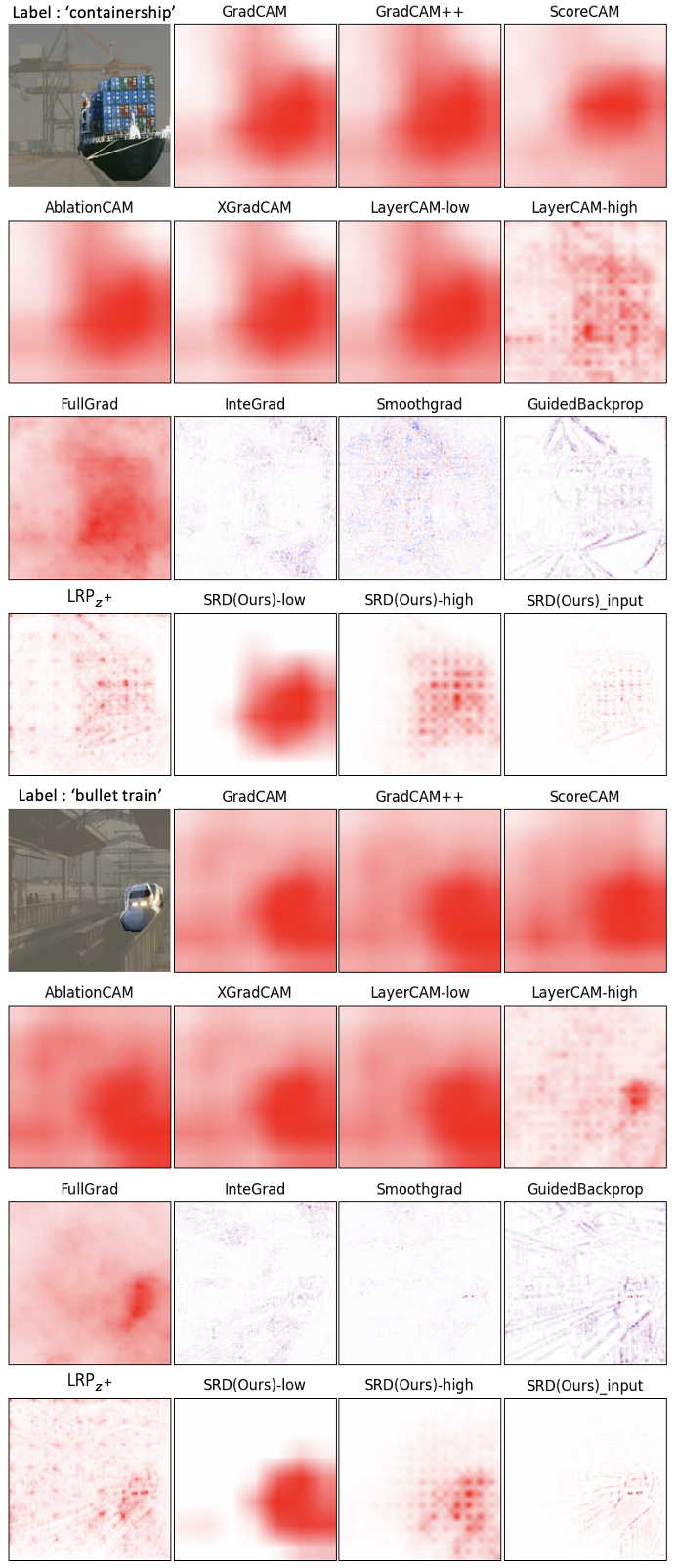}
\end{minipage}
\caption{Qualitative comparisons on ResNet50. The highlighted region denotes the segmentation mask.}
\label{fig:saliency_resnet1}
\end{figure*}

\begin{figure*}[t]
\centering
\begin{minipage}[t]{.495\textwidth}
  \centering
  \includegraphics[width=\linewidth]{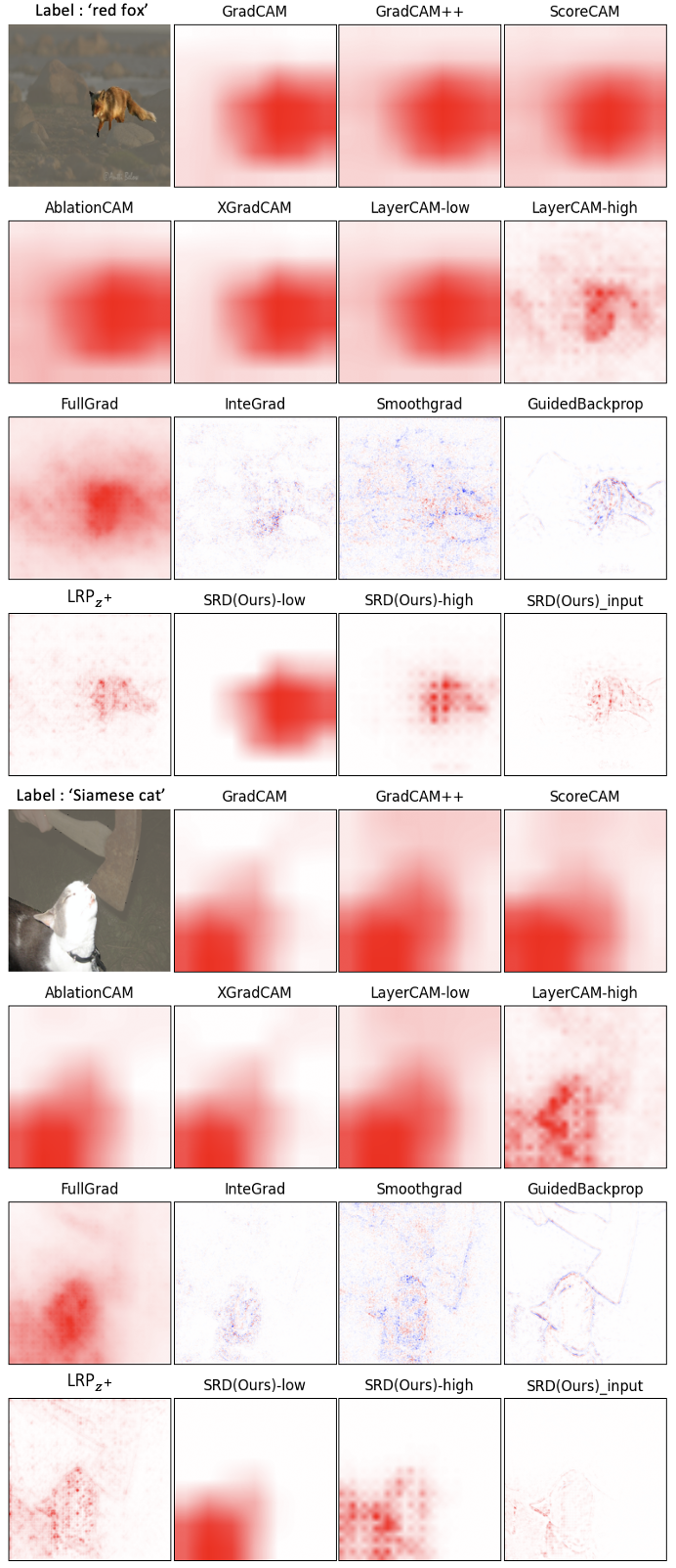}
\end{minipage}\hfill
\begin{minipage}[t]{.495\textwidth}
  \centering
  \includegraphics[width=\linewidth]{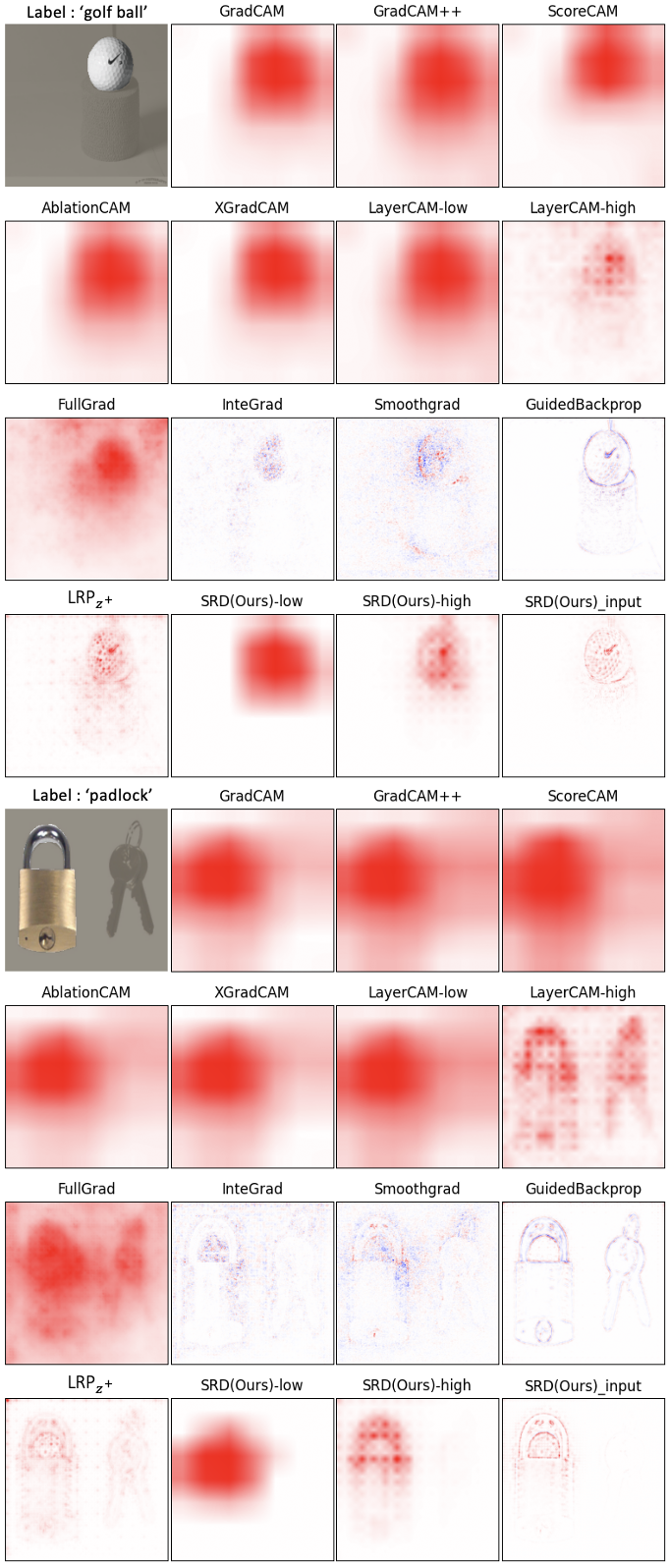}
\end{minipage}
\caption{Qualitative comparisons on ResNet50. The highlighted region denotes the segmentation mask.}
\label{fig:saliency_resnet2}
\end{figure*}

Fig.~\ref{fig:saliency_vgg1}-\ref{fig:saliency_resnet2} are some examples that compare the saliency maps of different methods.

\subsubsection{Explanation manipulation comparison}
\label{appendix: manipulation}
\begin{figure*}[t]
\centering
\begin{minipage}[t]{.495\textwidth}
  \centering
  \includegraphics[width=\linewidth]{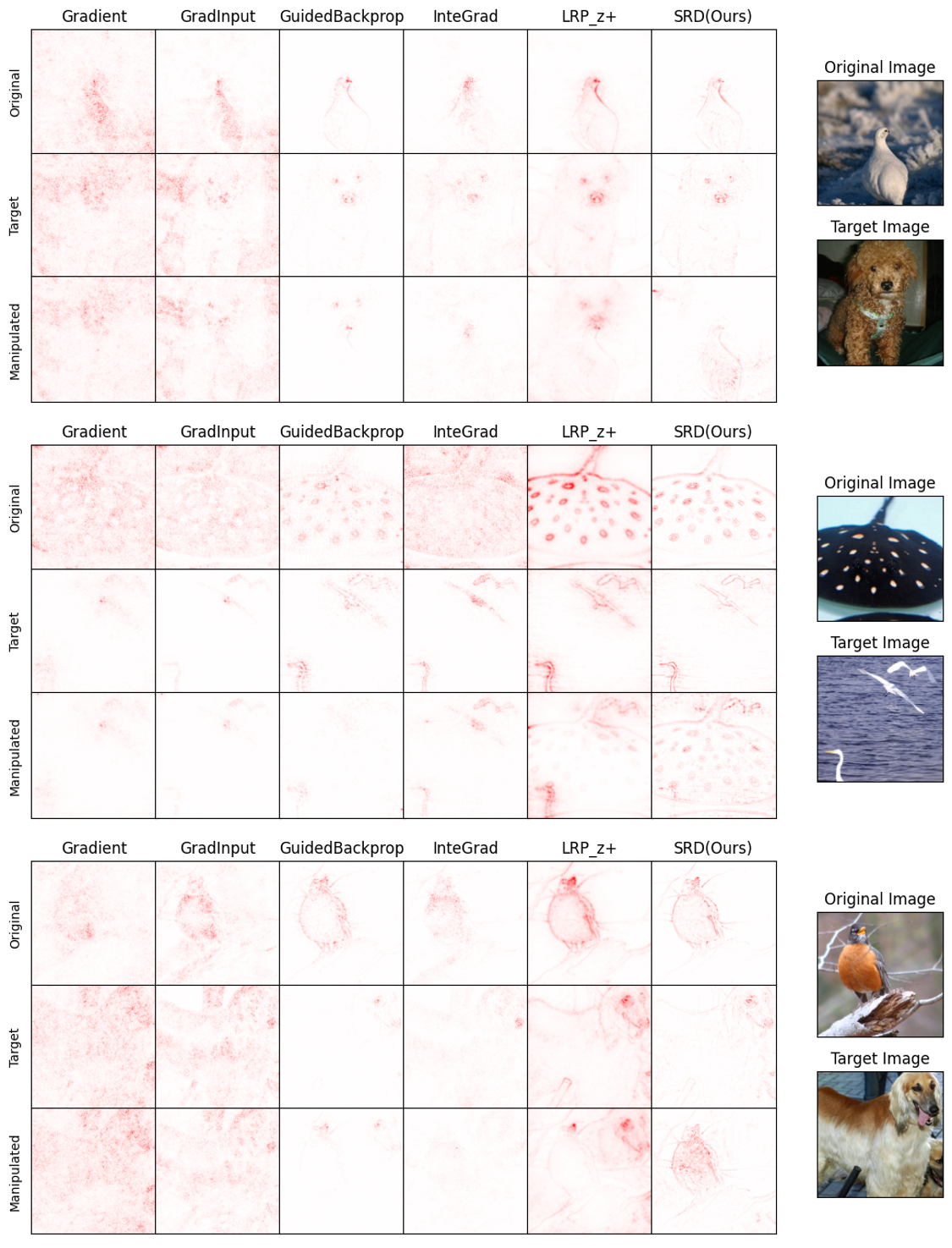}
\end{minipage}\hfill
\begin{minipage}[t]{.495\textwidth}
  \centering
  \includegraphics[width=\linewidth]{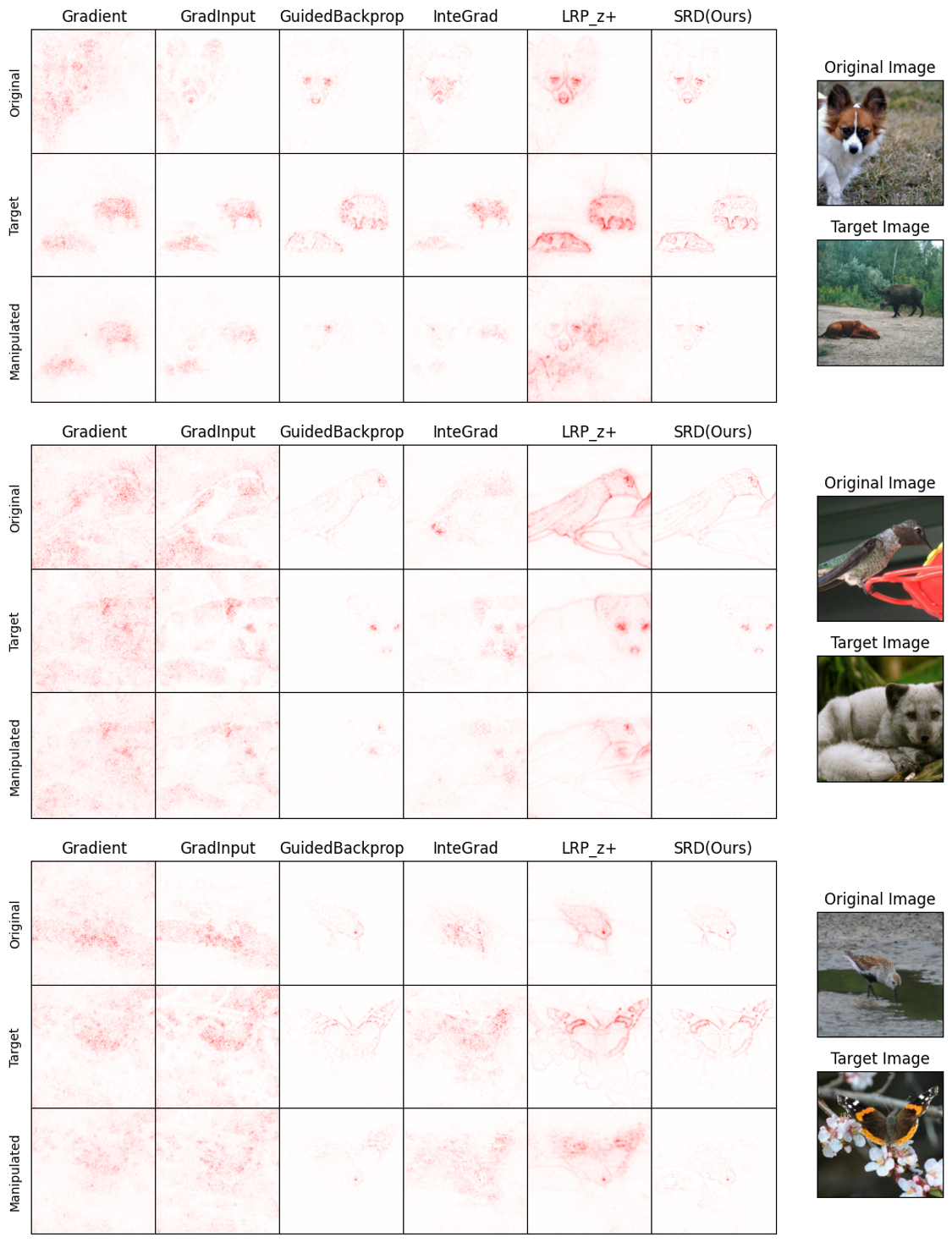}
\end{minipage}
\caption{Additional results on explanation manipulation comparison.}
\label{fig:manipulation1}
\end{figure*}

\begin{figure*}[t]
\centering
\begin{minipage}[t]{.495\textwidth}
  \centering
  \includegraphics[width=\linewidth]{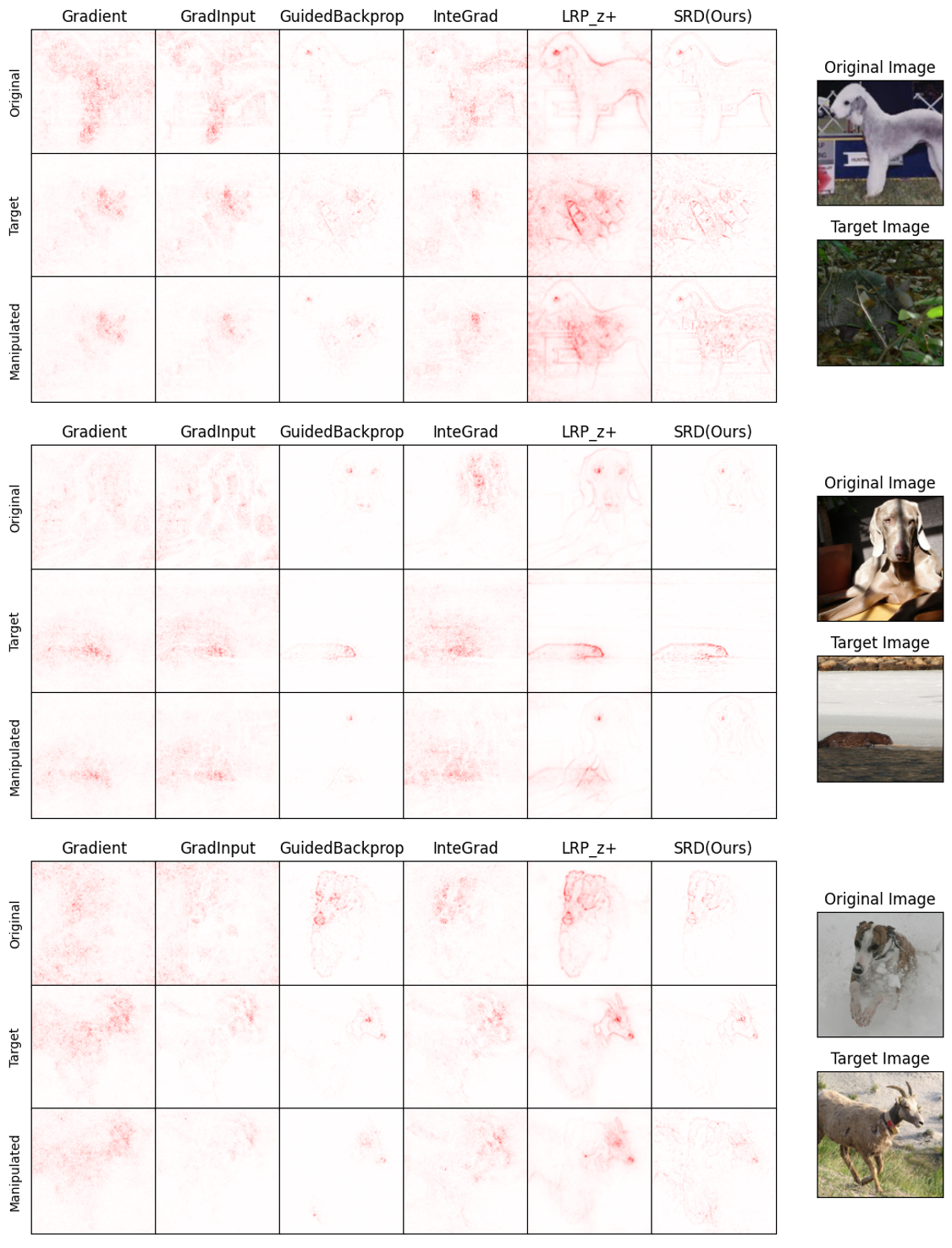}
\end{minipage}\hfill
\begin{minipage}[t]{.495\textwidth}
  \centering
  \includegraphics[width=\linewidth]{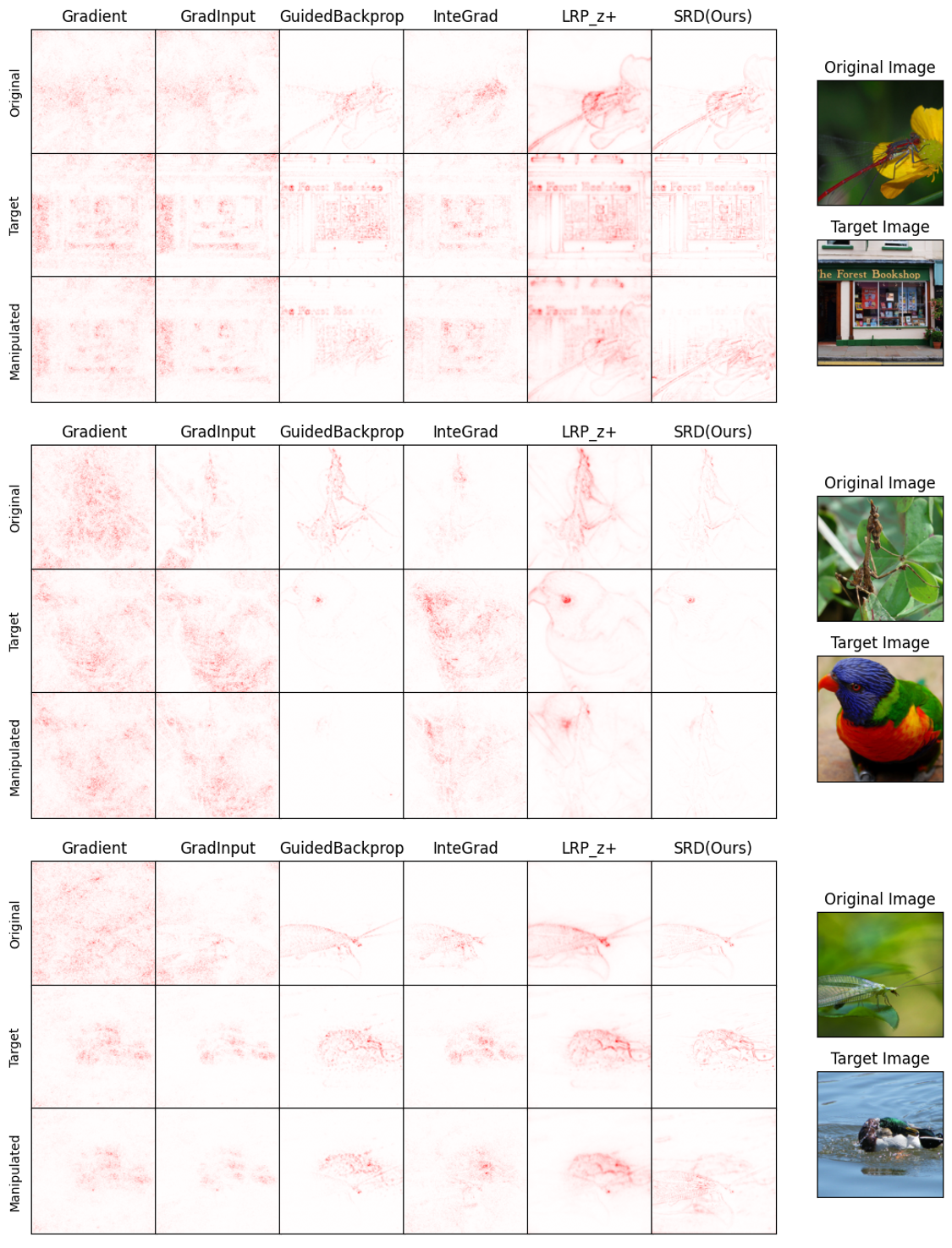}
\end{minipage}
\caption{Additional results on explanation manipulation comparison.}
\label{fig:manipulation2}
\end{figure*}

Fig.~\ref{fig:manipulation1} and Fig.~\ref{fig:manipulation2} are examples that compare explanation manipulation of different methods.

\subsubsection{Qulitative Result on application to various activations}
\label{appendix: qual_activation}

\yr{In this section, we evaluate the robustness of our proposed method, SRD (Sharing Ratio Decomposition), across different non-linear activation functions. 
While many attribution methods are sensitive to the specific type of non-linearity (e.g., ReLU) due to vanishing gradients or shattering gradients, our iERF-based approach maintains high-quality explanations.}

\begin{figure}[ht]
\begin{center}
\includegraphics[width=\linewidth]{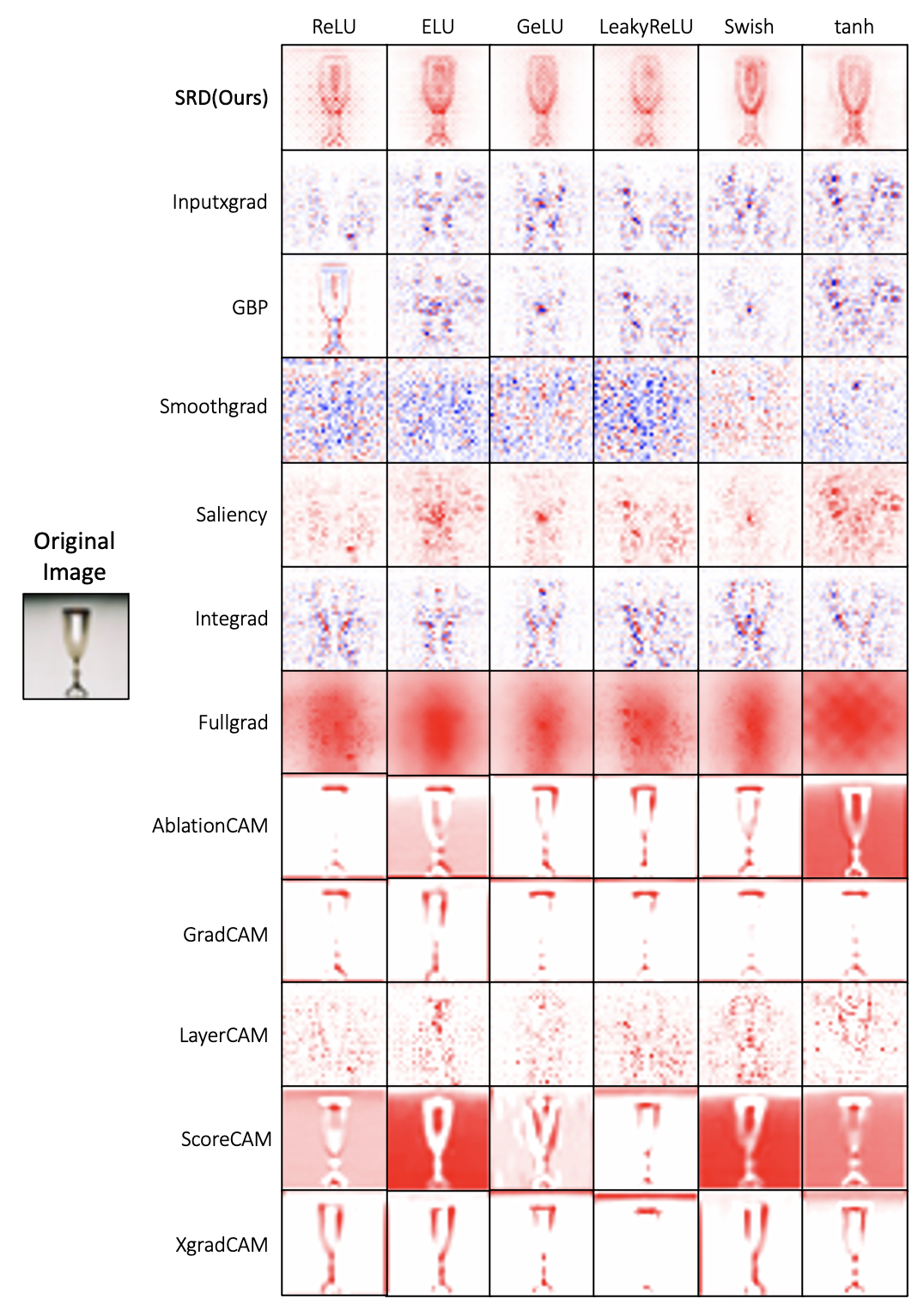}
\end{center}
\vspace{-5mm}
\caption{\textbf{Qualitative results across various activation functions.} Despite architectural changes in the nonlinearity, SRD continues to produce fine-grained and feasible explanation maps.}
\label{fig:activation}
\end{figure}

\subsection{Detail of metrics for local explanations}
\label{appendix: metrics}
\textbf{Pointing Game ($\uparrow$)} \cite{zhang2018top} evaluates the precision of attribution methods by assessing whether the highest attribution point is on the target.
The groundtruth region is expanded for some margin of tolerance (15px) to insure fair comparison between low-resolution saliency map and high-resolution saliency map.
Intuitively, the strongest attribution should be confined inside the target object, making a higher value for a more accurate explanation method. 
\begin{equation}
\mu_{\text{PG}} = \frac{Hits}{Hits+Misses}
\end{equation}

\textbf{Attribution Localization ($\uparrow$)} \cite{kohlbrenner2020towards} measures the accuracy of an attribution method by calculating the ratio , $\mu_{\text{AL}}$, between attributions located within the segmentation mask and the total attributions.
A high value indicates that the attribution method accurately explains the crucial features within the target object. 
\begin{equation}
\mu_{\text{AL}} = \frac{R_{in}}{R_{tot}},
\end{equation}
where $\mu_{\text{AL}}$ is an inside-total relevance ratio without consideration of the object size. 
$R_{in}$ is the sum of positive relevance in the bounding box, $R_{tot}$ is the total sum of positive relevance in the image.

\textbf{Sparseness ($\uparrow$)} \cite{chalasani2020concise} evaluates the density of the attribution map using the Gini index.
A low value indicates that the attribution is less sparse, which may be observed in low-resolution or noisy attribution maps. 
\begin{equation}
    \mu_{\text{Spa}} = 1 - 2\sum^d_{k=1}\frac{v_{(k)}}{||\mathbf{v}||_1}(\frac{d - k + 0.5}{d}),
\end{equation}
where $\mathbf{v}$ is a flatten vector of the saliency map $\phi(x)$

\textbf{Fidelity ($\uparrow$)} \cite{ijcai2020p417} measures the correlation between classification logit and attributions. 
Randomly selected 200 pixels are replaced to value of 0. 
The metric then measures the correlation between the drop in target logit and the sum of attributions for the selected pixels.
\[
\mu_{\text{Fid}}= \underset{S \in \binom{[d]}{|S|}}{\text{Corr}}\left(\sum_{i \in S}\phi(x)_i, F(x) - F\left(x_{[x_{s} = \bar{x}_{s}]}\right)\right),
\]
where $F$ is the classifier, $\phi(x)$ the saliency map given $x$

\textbf{Stability $(\downarrow)$} \cite{alvarez2018towards} evaluates the stability of an explanation against noise perturbation.
While measuring robustness against targeted perturbation can be computationally intensive and complicated due to non-continuity of some attribution methods, a weaker robustness metric is introduced to assess stability against random small perturbations.
This metric calculates the maximum distance between the original attribution and the perturbed attribution for finite samples.
A low stability score is preferred, indicating a consistent explanation under perturbation. 
\begin{equation}
\mu_{\text{Sta}} = \max_{x_j \in \mathit{N}_\epsilon(x_i)} \frac{\| \phi(x_i) - \phi(x_j) \|_2}{ \| x_i - x_j \|_2},
\end{equation}
where $\mathit{N}_\epsilon(x_i)$ is a gaussian noise with standard deviation 0.1. 
all of the metrics are measure after clamping the attributions to [-1,1], as all the attrubution methods are visualized after clamping.

\subsection{All-layer Analysis}
\label{appendix: mechanistic_all}
\begin{figure*}[ht!]
    \centering
    \includegraphics[width=.9\linewidth]{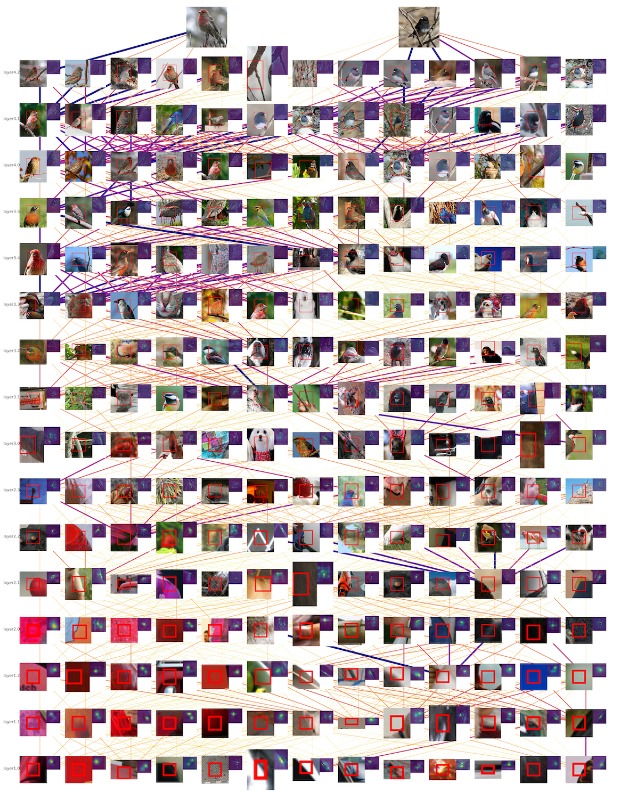}
    \caption{Mechanistic concept explanation graph of every layer in ResNet50. The top-5 most important concepts in each class and top-3 shared concepts. The thicker and bluer the edge, the stronger the contribution between concepts.}
    \label{fig:appendix_main}
\end{figure*}
\begin{figure*}[ht!]
    \centering
    \includegraphics[width=.9\linewidth]{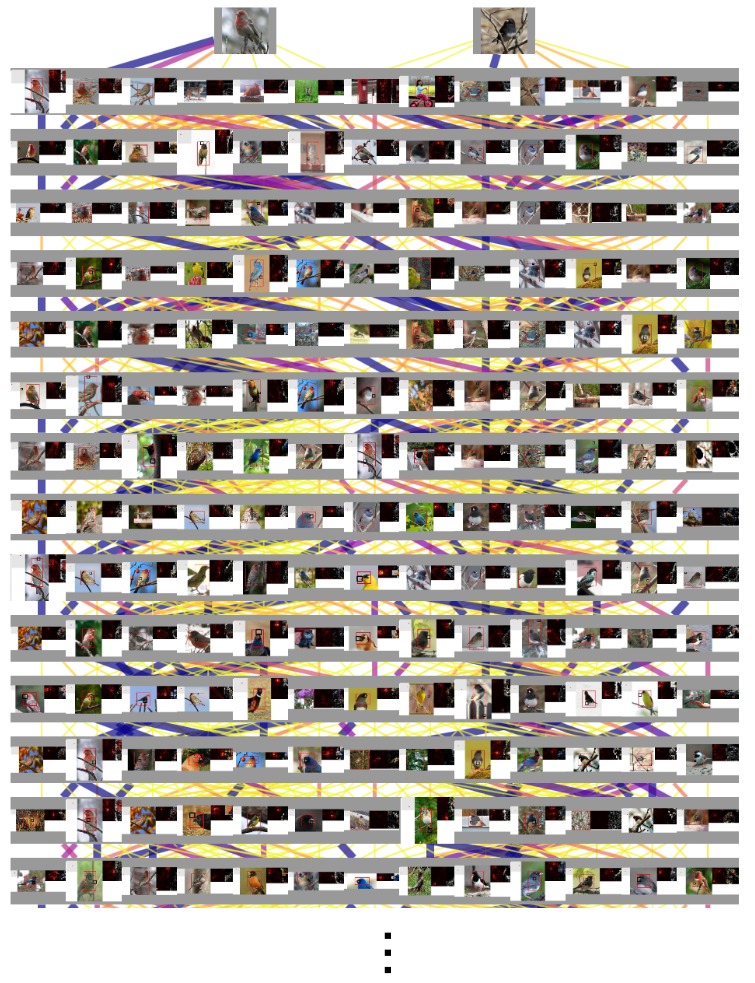}
    \caption{Mechanistic concept explanation graph of every layer in VIT-g-14. The top-5 most important concepts in each class and top-3 shared concepts. The thicker and bluer the edge, the stronger the contribution between concepts.}
    \label{fig:appendix_main_vit}
\end{figure*}
\begin{figure*}[ht!]
    \centering
    \includegraphics[width=.9\linewidth]{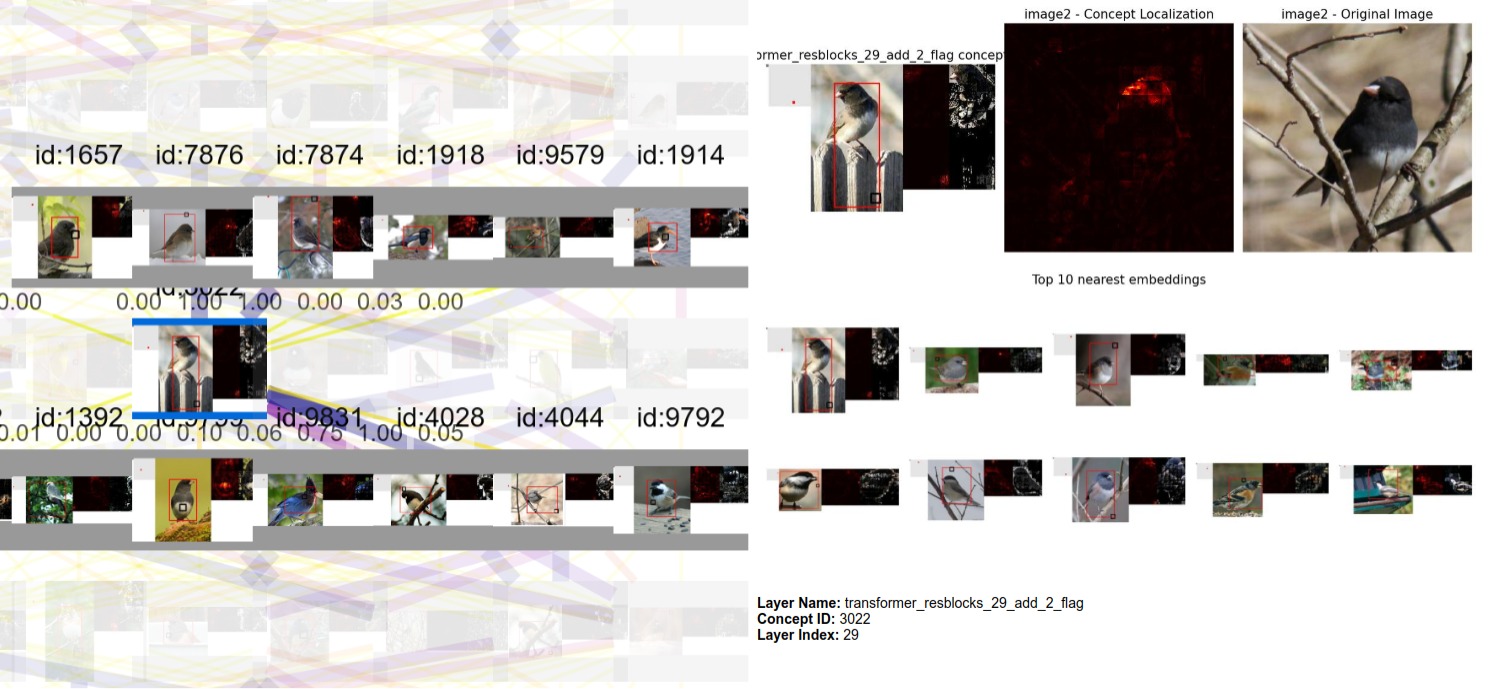}
    \caption{An example of graph of concept 3022 at 29th layer of VIT-g-14. Left: the connection between concepts. Right: the explanation of selected concept. It shows the iERF of selected concept for the input, with top 10 nearest PFV across the dataset.}
    \label{fig:vit_example}
\end{figure*}
As in Fig.~\ref{fig:appendix_main}, ~\ref{fig:appendix_main_vit},~\ref{fig:vit_example}, we can decompose every concepts through every layers of Resnet50 and VIT-g-14. 
The ICAT scores between layers are normalized to [0,1].

\subsection{\ssy{A Note on PFV-level Attribution from Neuron-wise Relevance}}
\label{app:pfv_neuronwise_note}

This note clarifies a unit-mismatch that arises when the explanatory target is a pointwise feature vector (PFV), while the attribution method assigns relevance to individual neurons.

Let $v_i^l \in \mathbb{R}^{C_l}$ denote the source PFV at spatial location $i$ in layer $l$, and let $v_j^k \in \mathbb{R}^{C_k}$ denote the target PFV at spatial location $j$ in layer $k$, with $l<k$. 
Suppose a neuron-wise propagation method yields channel-wise relevance scores
\[
r_{i\to j}^{\,l\to k}
=
\bigl(r_{i\to j,1}^{\,l\to k},\dots,r_{i\to j,C_k}^{\,l\to k}\bigr)
\in \mathbb{R}^{C_k},
\]
where $r_{i\to j,c}^{\,l\to k}$ denotes the relevance assigned to the contribution of source location $i$ toward the $c$-th channel of the target PFV at location $j$.

If one wishes to summarize these channel-wise scores as a single PFV-level quantity, an additional aggregation map is required:
\[
R_{i\to j}^{\mathrm{PFV}}
=
G\!\left(r_{i\to j}^{\,l\to k}\right),
\qquad
G:\mathbb{R}^{C_k}\to\mathbb{R}.
\]
However, the propagation rule itself does not canonically specify the choice of $G$. 
For example, one may consider the signed sum,
\[
G_{\mathrm{sum}}(r)=\sum_{c=1}^{C_k} r_c,
\]
the absolute sum,
\[
G_{\mathrm{abs}}(r)=\sum_{c=1}^{C_k} |r_c|,
\]
or the positive-part sum,
\[
G_{+}(r)=\sum_{c=1}^{C_k} \max(r_c,0).
\]
These choices need not agree, and can lead to different PFV-level scores.

A simple example illustrates the point. Consider two candidate source PFVs with channel-wise relevance vectors
\[
r^{(a)}=(1,-1),
\qquad
r^{(b)}=(0.4,0.4).
\]
Under signed aggregation,
\[
G_{\mathrm{sum}}\!\left(r^{(a)}\right)=0,
\qquad
G_{\mathrm{sum}}\!\left(r^{(b)}\right)=0.8,
\]
so one obtains
\[
R^{\mathrm{PFV}}(a) < R^{\mathrm{PFV}}(b).
\]
In contrast, under absolute aggregation,
\[
G_{\mathrm{abs}}\!\left(r^{(a)}\right)=2,
\qquad
G_{\mathrm{abs}}\!\left(r^{(b)}\right)=0.8,
\]
which reverses the ordering:
\[
R^{\mathrm{PFV}}(a) > R^{\mathrm{PFV}}(b).
\]
Likewise, under positive-part aggregation,
\[
G_{+}\!\left(r^{(a)}\right)=1,
\qquad
G_{+}\!\left(r^{(b)}\right)=0.8,
\]
again yielding
\[
R^{\mathrm{PFV}}(a) > R^{\mathrm{PFV}}(b).
\]

Thus, when the explanatory target is a PFV, neuron-wise relevance does not by itself yield a canonical PFV-level score; an extra aggregation across channels is required. This is not a limitation of neuron-wise relevance for neuron-wise questions. Rather, it reflects a mismatch between the attribution unit and the representation unit in our setting.

SRD avoids this mismatch by defining contribution directly at the PFV level through the sharing ratio in Eq.~(3). In other words, SRD does not first assign channel-wise relevance and then aggregate it afterward into a PFV-level quantity; the PFV itself is the primitive attribution unit from the outset.

\end{document}